\documentclass{article}

\usepackage[preprint]{neurips_2026}

\usepackage[utf8]{inputenc} % allow utf-8 input
\usepackage[T1]{fontenc}    % use 8-bit T1 fonts
\usepackage{hyperref}       % hyperlinks
\usepackage{url}            % simple URL typesetting
\usepackage{booktabs}       % professional-quality tables
\usepackage{amsfonts}       % blackboard math symbols
\usepackage{amsmath}        % math text environments (\text, \mathrm, etc.)
\usepackage{nicefrac}       % compact symbols for 1/2, etc.
\usepackage{microtype}      % microtypography
\usepackage{xcolor}         % colors
\usepackage{graphicx}       % figures (\includegraphics)

\title{Hierarchical Self-Improvement: A Framework for Task-Specific Evolvable Agent Harnesses}

\author{%
  Tailin Zhou \\
  HKUST \\
  \texttt{tzhouaq@connect.ust.hk} \\
}

\begin{document}

\maketitle

\begin{abstract}

Modern LLM agents are often improved by modifying prompts, tools, or workflows manually, while the executable scaffold surrounding the model---the \emph{harness}---is typically treated as a fixed artifact after deployment. 
 This work studies an alternative where the harness is \emph{task-specific and continuously evolvable}: each task family maintains its own harness, which is hot-swapped across iterations through a fixed task-injection seam and rewritten using environment feedback. We introduce \textbf{Hierarchical Self-Improvement (HSI)}, a framework in which a single frozen LLM $M$ operates across three hierarchical scopes: a task harness $H$ that executes tasks, an evolver that rewrites $H$, and a meta-evolver that rewrites the evolver's strategy code under a frozen outer anchor. A thinking-on/off design isolates the contribution of harness evolution by disabling reasoning during task execution while enabling it during self-modification. HSI is bounded by two factors: a \emph{feedback-fidelity bound}, since evolution requires informative reward signals to guide selection, and a \emph{backbone capability bound}, since harness redesign cannot overcome limitations of the frozen model. On BALROG with DeepSeek-V4-Flash-Preview as the frozen backbone, HSI achieves consistent gains over the initial harness on moderate-difficulty tasks ($+39.3$ on BabyAI, $+33.0$ on Crafter, $+25.0$ on TextWorld, and $+15.0$ on MiniHack, all in raw \% Progress), while obtaining strong held-out generalization on BabaIsAI sub-suites ($0.98$ best-test on BreakStop and $1.00$ on GoTo from a $20\%$ unseen split). On tasks beyond the backbone's capability (NLE), harness evolution provides no improvement. These results demonstrate task-specific harness evolution as a viable axis for improving frozen LLM agents under clear empirical limits. Code is available at \url{https://github.com/TailinZhou/hsi}.

\end{abstract}

\section{Introduction}

The executable scaffold surrounding an LLM (the \emph{harness}, including prompts, tool orchestration, memory, and verification logic) has emerged as a critical determinant of agent performance, with different harness designs producing substantial gaps even under identical model backbones~\citep{lee2026metaharnessendtoendoptimizationmodel,yao2026harnessbenchmeasuringharnesseffects}. However, two fundamental challenges remain unresolved. First, existing self-improvement approaches have not fully reached the harness layer endogenously. G\"{o}del-style self-improvement systems primarily evolve the agent's per-step decision code~\citep{yin2025godelagentselfreferentialagent,zhang2026darwingodelmachineopenended,wang2025huxleygodelmachinehumanlevelcoding,weng2026groupevolvingagentsopenendedselfimprovement,zhang2026hyperagents}, while harness-engineering approaches that target broader scaffolds often rely on external proposers or stronger designer models~\citep{lee2026metaharnessendtoendoptimizationmodel,zhang2026selfharnessharnessesimprove,lin2026agenticharnessengineeringobservabilitydriven}. Second, it remains unclear whether observed harness-evolution gains reflect genuine capability improvement or merely test-time search, as recent evaluations reveal substantial overfitting and limited generalization~\citep{wang2026rethinkingevaluationharnessevolution}. These limitations raise a central question: \emph{when the underlying model is frozen, can an agent endogenously evolve its own harness to improve performance, and what ultimately limits such improvement?}

We address this question with \textbf{Hierarchical Self-Improvement (HSI)}, a framework that enables a single frozen LLM $M$ to improve through hierarchical evolution of its own harness rather than parameter updates. HSI operates across three layered scopes: the \emph{task harness} $H$, which executes tasks around $M$; the \emph{evolver}, which rewrites $H$ across iterations; and the \emph{meta-evolver}, which rewrites the evolver's strategy code one layer above. To prevent unrestricted self-reference, the meta-evolver's own execution logic remains frozen as an outer anchor, restricting self-modification to layered and empirically validated edits. A thinking-on/off design isolates the contribution of harness evolution: reasoning is disabled during task execution to fix the model's per-step capability ceiling and enabled during rewriting to maximize the chance of successful self-modification. The resulting loop consists of five stages: seed selection, main evolution, commit selection, meta-evolution, and terminal best-version selection.

We instantiate HSI on BALROG~\citep{paglieri2025balrog}, a benchmark of long-horizon text-based interactive games exhibiting a natural difficulty gradient. Using DeepSeek-V4-Flash-Preview as the frozen backbone, we evaluate harness evolution under two complementary protocols: full-set evolution to measure achievable in-distribution improvement and split evolution to assess held-out generalization. Across moderate-difficulty environments including TextWorld, Crafter, and BabyAI, HSI consistently improves over matched initial harness baselines while keeping the backbone and inference configuration fixed. On easier BabaIsAI sub-suites, evolved harnesses generalize to unseen tasks, while on NLE, where the backbone provides insufficient capability and feedback signal, harness evolution yields no meaningful improvement. These results reveal both the potential and the limits of endogenous harness evolution for frozen LLM agents.

HSI instantiates a general template of \emph{task-specific and continuously evolvable harnesses}: each task family maintains its own harness, which can be hot-swapped across iterations through a fixed task-injection seam and refined using environment feedback. The key design principle is hierarchical evolution under frozen anchors: the task harness changes while the rewriting procedure remains anchored, and the rewriting procedure itself can evolve one level above while the outer execution anchor remains fixed. By keeping the same frozen $M$ across all scopes, HSI realizes endogenous harness evolution without unrestricted self-reference. This paradigm is subject to two fundamental limits: a \emph{feedback-fidelity bound}, since evolution requires informative reward signals to guide selection, and a \emph{backbone capability bound}, since harness redesign cannot overcome limitations of the underlying model.
Our contributions are:
\begin{enumerate}
    \item \textbf{HSI framework.} 
    We introduce a hierarchical self-improvement framework that enables a frozen LLM to evolve its own task harness through nested rewriting scopes, while preventing unrestricted self-reference through a frozen outer anchor.

    \item \textbf{Positive evidence under a controlled model ceiling.}
    On BALROG with a frozen DeepSeek-V4-Flash backbone, we show that task-specific harness evolution achieves consistent gains on moderate-difficulty environments and generalizes to held-out BabaIsAI sub-suites. The thinking-off task execution protocol isolates inference-time reasoning as a confounding factor.

    \item \textbf{Empirical characterization of scaling limits.}
    We identify two practical boundaries of harness evolution: feedback availability and backbone capability. Across environments with different difficulty levels, these boundaries characterize when endogenous harness evolution can improve frozen LLM agents and when further redesign provides limited benefit.
\end{enumerate}

\section{Related Work}

HSI relates to three areas of research: G\"odel-style self-improvement, harness engineering for LLM agents, and the evaluation and theoretical characterization of self-improving systems. We review these directions and clarify how HSI differs from existing approaches.

\subsection{G\"odel-Style Self-Improvement}

The G\"odel Machine~\citep{schmidhuber2003godelmachine} introduced the foundational idea of a self-improving system that can modify its own program, including the procedure responsible for future modifications, provided that such changes can be verified to improve performance. Recent works have operationalized this idea for LLM agents through runtime self-editing, evolutionary search, and population-based exploration. G\"odel Agent~\citep{yin2025godelagentselfreferentialagent} realizes self-referential improvement through runtime code modification, while Darwin G\"odel Machine (DGM)~\citep{zhang2026darwingodelmachineopenended}, Huxley-G\"odel Machine (HGM)~\citep{wang2025huxleygodelmachinehumanlevelcoding}, and Group-Evolving Agents (GEA)~\citep{weng2026groupevolvingagentsopenendedselfimprovement} explore increasingly broader evolutionary mechanisms. HyperAgents~\citep{zhang2026hyperagents} further extends this direction by allowing meta-level mechanisms to become editable. However, these approaches primarily place the evolvable boundary at the agent's decision procedure or program execution process, leaving open whether the broader agent harness can be evolved endogenously.

\subsection{Harness Engineering and Online Adaptation}

Harness engineering studies how the executable components surrounding an LLM, including prompts, tools, memory, and verification mechanisms, shape agent behavior and performance. Meta-Harness~\citep{lee2026metaharnessendtoendoptimizationmodel} formalized harness optimization as an outer-loop search problem using stronger proposer models, while AutoHarness~\citep{lou2026autoharnessimprovingllmagents} explored automatic harness synthesis through search-based optimization. Recent efforts have investigated endogenous harness adaptation, where agents modify their own scaffolds through feedback and validation. Self-Harness~\citep{zhang2026selfharnessharnessesimprove} studies self-generated harness refinement through weakness mining and regression-based acceptance, while Agentic Harness Engineering (AHE)~\citep{lin2026agenticharnessengineeringobservabilitydriven} identifies observability and component-level feedback as key factors for effective self-improvement. HarnessX~\citep{chen2026harnessxcomposableadaptiveevolvable} further formalizes harnesses as composable adaptive systems, and HarnessForge~\citep{chen2026harnessforgejointharnesspolicy} studies joint harness-policy evolution.

A parallel line of work considers harness adaptation during deployment. TTHE~\citep{nie2026tthetesttimeharnessevolution} evolves harnesses using execution traces at test time, while Live-SWE-Agent~\citep{xia2025livesweagentsoftwareengineeringagents}, Continual Harness~\citep{karten2026continualharnessonlineadaptation}, and Adaptive Auto-Harness~\citep{liu2026adaptiveautoharnesssustainedselfimprovement} investigate online and continual adaptation. Despite these advances, most existing approaches focus on evolving the task harness itself, rather than evolving the mechanism that governs how harnesses are discovered, selected, and rewritten.

\subsection{Evaluation, Attribution, and Theoretical Limits}

Recent work has highlighted the difficulty of evaluating self-improving agents and attributing gains to genuine capability improvement. Harness-Bench~\citep{yao2026harnessbenchmeasuringharnesseffects} establishes harness design as an independent evaluation axis, showing substantial performance variation across harness configurations under identical models. Harness Updating Is Not Harness Benefit~\citep{lin2026harnessupdatingharnessbenefit} further separates an agent's ability to produce harness updates from its ability to benefit from those updates. Rethinking the Evaluation of Harness Evolution~\citep{wang2026rethinkingevaluationharnessevolution} shows that automatic harness evolution may fail to outperform simple test-time scaling baselines and can suffer from severe overfitting.

From a theoretical perspective, \citet{wang2026statisticallimitsselfimprovingagents} characterize statistical limits of self-improving agents and show that distribution-free PAC guarantees are preserved under self-modification when the reachable hypothesis family has bounded complexity. This provides a theoretical perspective for understanding capability boundaries in self-improving systems.

\subsection{Comparison with Prior Work}

Rather than resurvey existing approaches, we organize the comparison around three design questions that motivate HSI.

\paragraph{Where does the evolvable boundary lie?}

G\"odel-style self-improvement approaches~\citep{yin2025godelagentselfreferentialagent,zhang2026darwingodelmachineopenended,wang2025huxleygodelmachinehumanlevelcoding,weng2026groupevolvingagentsopenendedselfimprovement,zhang2026hyperagents} primarily evolve the agent's decision procedure or execution code. Harness-engineering approaches~\citep{lee2026metaharnessendtoendoptimizationmodel,zhang2026selfharnessharnessesimprove,lin2026agenticharnessengineeringobservabilitydriven} target a broader scaffold but often rely on external optimization signals or stronger proposer models. HSI studies a broader endogenous editable surface: the harness that coordinates prompts, tools, memory, state, and cross-step interaction patterns, while the same frozen model that executes tasks also performs the evolution.

\paragraph{How should harness evolution scale?}

A common assumption in prior harness optimization is that a single improved harness can generalize across tasks. However, recent evaluation studies reveal that evolved harnesses may overfit and fail to outperform simple test-time scaling baselines~\citep{wang2026rethinkingevaluationharnessevolution}. HSI instead adopts a task-specific evolution paradigm, where each task family maintains its own evolving harness and generalization is evaluated through held-out task splits. The scaling axis is therefore not a universal harness optimized once, but continuous evolution of task-specific harnesses under controlled evaluation.

\paragraph{What should remain fixed during self-modification?}

Prior analyses argue that self-improving systems require explicit constraints on editable surfaces and external evaluation mechanisms~\citep{weng2026harness,wang2026statisticallimitsselfimprovingagents}. HSI adopts a hierarchical constraint: the task harness is hot-swappable through a fixed \texttt{using\_harness} interface, the evolver strategy is editable at a higher level, and the meta-evolver operates under a frozen outer anchor. In addition, evaluation signals and data splits remain outside the agent's control. This design provides an empirical framework for studying endogenous harness evolution while maintaining clear boundaries on self-modification.

\section{Hierarchical Self-Improvement}

We present HSI, a framework in which a single frozen LLM improves task performance through layered self-modification of its harness and the procedure that evolves it. HSI separates the evolving task-facing scaffold from the evolution mechanism itself through a hierarchical architecture. We first introduce the design principles that define this setting (\S\ref{sec:design}), then present the overall architecture and evolution loop (\S\ref{sec:framework}), followed by the detailed stages and their structural invariants (\S\ref{sec:stages}).

\subsection{Design Principles}
\label{sec:design}

In HSI, the harness is \emph{task-specific and continuously evolvable}: each task family maintains its own harness, which is hot-swappable across iterations through a fixed task-injection seam and refined using environment feedback. This design naturally requires hierarchical separation. A single-layer design would couple the task-facing behavior of the harness with the strategy responsible for rewriting it, making the object of optimization and the optimizer itself inseparable. Hierarchical separation instead allows the task harness to evolve while its rewriting procedure remains anchored, and allows the rewriting procedure to evolve one layer above while preserving a frozen outer anchor. This localizes self-modification to structured and empirically validated edits rather than unrestricted self-reference.

The following two principles define how hierarchical self-modification is realized under a frozen LLM $M$. Throughout this work, the underlying model parameters remain fixed; improvements arise solely from modifications to the harness and the procedure that governs its evolution.

\paragraph{Principle 1 (Single frozen model, three harness scopes).}
HSI operates with a single frozen LLM $M$ across three hierarchical scopes. In the \emph{task-harness scope}, $M$ executes a task harness $H$ to interact with the environment. In the \emph{evolver scope}, $M$ modifies $H$ across iterations through seed selection, harness evolution, and candidate selection. In the \emph{meta-evolver scope}, $M$ modifies the evolver strategy itself, including decisions such as seed generation, commit selection, archive maintenance, and final version export.

All three scopes share the same frozen model $M$, prompt format, and \texttt{react()} primitive. They differ only in available tools and execution context. The scopes are separated by explicit memory boundaries: task-harness, evolver, and meta-evolver interactions maintain independent histories rather than representing independent agents. Unlike external-proposer approaches~\citep{zhang2026selfharnessharnessesimprove,lin2026agenticharnessengineeringobservabilitydriven,lee2026metaharnessendtoendoptimizationmodel}, HSI uses the same frozen model for both execution and evolution.

\paragraph{Principle 2 (Self-determined explore--exploit).}
HSI does not prescribe an explicit explore--exploit schedule during evolution. The framework does not determine when the agent should inspect code, evaluate candidates, commit changes, record lessons, or allocate effort between exploration and exploitation. Instead, these decisions are treated as part of the evolvable strategy controlled by $M$.

The framework only provides atomic interaction primitives, evolutionary feedback signals, and structural invariants. These include evaluation rewards, committed-version lineage, and persistent lessons stored in \texttt{BOOTSTRAP.md}. The agent is responsible for determining how these signals should be combined into an effective evolution strategy, while the framework constrains only the boundaries within which such adaptation occurs.

\subsection{Hierarchical Architecture and Self-Governing Agent Loop}
\label{sec:framework}

\begin{figure}[t]
\centering
\includegraphics[width=0.93\linewidth]{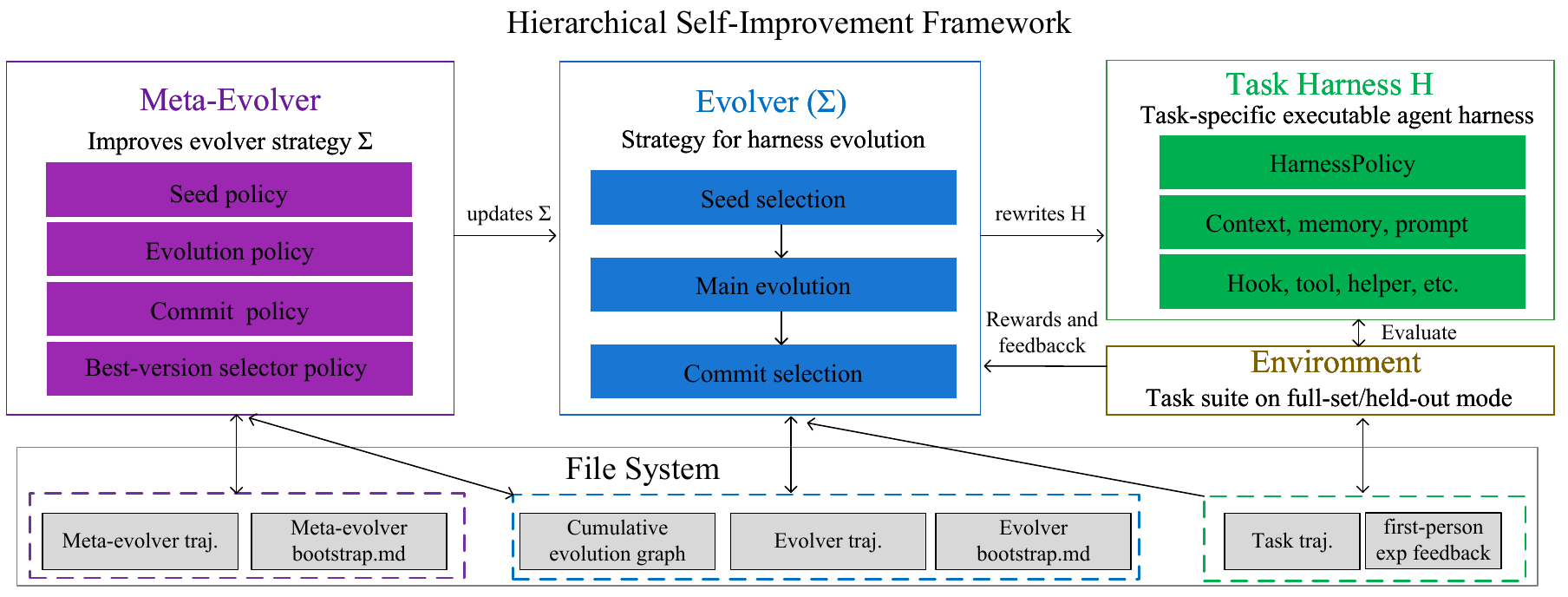}
\caption{The HSI framework. A single frozen LLM $M$ operates across three hierarchical scopes with disjoint editable surfaces. The task-harness scope executes the task-specific harness $H$; the evolver scope rewrites $H$ through seed selection, main evolution, and commit selection; and the meta-evolver scope rewrites the evolver strategy $\Sigma$, including seed, evolution, commit, and final version selection policies.}
\label{fig:framework}
\end{figure}

The design principles above define an architecture in which a single frozen LLM $M$ operates across multiple hierarchical scopes with different editable surfaces. The central idea is recursive but bounded self-modification: the task harness $H$ can evolve, the procedure that evolves $H$ (the evolver strategy $\Sigma$) can itself evolve, while the outer execution logic that governs $\Sigma$ remains frozen. This separation allows HSI to improve both the object of optimization and the optimization procedure without introducing unrestricted self-reference.

\paragraph{Three hierarchical scopes.}

HSI consists of three scopes operated by the same frozen model $M$ (Figure~\ref{fig:framework}). The \emph{task-harness scope} executes the current task harness $H$, which contains the task-facing components of the agent, including prompts, tools, memory, state management, hooks, and the policy that determines how the model interacts with the environment. The harness is the primary editable object during evolution and is connected to tasks through a fixed injection seam.

The \emph{evolver scope} modifies the task harness across iterations. It executes the evolution procedure defined by $\Sigma$, which determines how candidate harnesses are generated, evaluated, and selected. Each iteration consists of three stages: seed selection, main evolution, and commit selection. The evolver receives environment feedback through harness evaluation and maintains persistent evolutionary information through the evolution archive and bootstrap memory.

The \emph{meta-evolver scope} operates one level above the harness evolution process. It modifies $\Sigma$ itself, including the strategies responsible for seed generation, candidate evolution, commit selection, and final version selection. Unlike the task harness and evolver strategy, the execution logic that performs meta-evolution is not editable. It is loaded from an immutable initialization template and serves as the outer frozen anchor that bounds self-modification.

\paragraph{Self-governing evolution loop.}

The three scopes share the same frozen model $M$, prompt format, and reasoning primitive, but maintain separate execution contexts and memory boundaries. The complete evolution process forms a cyclic loop consisting of four iterative stages: seed selection, main evolution, commit selection, and meta-evolution. The first three stages modify the task harness $H$, while the fourth stage modifies the evolution strategy $\Sigma$. After $T$ iterations, a terminal best-version selection stage exports the final harness for evaluation.

Unlike predefined optimization pipelines, HSI does not specify an explicit exploration schedule within these stages. The same frozen model determines how to inspect, modify, evaluate, and retain candidate versions based on the available tools and feedback signals. The framework only provides structural constraints, including the editable boundaries, evaluation interface, and frozen outer anchor. This design enables endogenous evolution of both the harness and its optimization strategy while preserving a controlled self-improvement boundary.

\subsection{Stage Mechanisms}
\label{sec:stages}

The evolution process consists of five stages. The first three stages operate on the task harness $H$, the fourth stage modifies the evolution strategy $\Sigma$, and the final stage selects the exported harness for evaluation. Each stage is implemented as a bounded \texttt{react()} loop whose central decisions are delegated to the frozen LLM $M$, while the editable surface and structural invariants are explicitly constrained. See Appendix \ref{app:implementation} for implementation  details.

\subsubsection{Seed Selection: Hypothesis Generation}
\label{sec:select-seed}

Let $H_t$ denote the harness at the beginning of iteration $t$, and let $\mathcal{G}_t=(V_t,E_t)$ denote the cumulative evolution graph. Each node $v\in V_t$ corresponds to a committed harness snapshot annotated with reward $r_v$, while edges store semantic relations describing how one version relates to another.

Seed selection chooses an ancestor from $\mathcal{G}_t$ and generates a structured hypothesis for the next evolution iteration:
\begin{equation}
(\hat{H}_t,h_t)=\mathrm{seed\_selection}(\mathcal{G}_t,M).
\end{equation}

The decision is made by $M$ using previous rewards, evolution history, and accumulated lessons. The generated hypothesis contains four elements: the selected anchor version, the motivation for choosing it, the expected improvement direction, and a falsification criterion. This hypothesis is injected into the subsequent evolution process, transforming evolution from unconstrained mutation into a goal-directed search guided by explicit predictions.

\subsubsection{Main Evolution: Hot-Swappable Harness Rewrites}
\label{sec:main-evolve}

Main evolution is the stage where $M$ directly modifies the task harness $H$. Given the seeded harness $\hat{H}_t$ and hypothesis $h_t$, the evolver generates candidate versions:

\begin{equation}
\{V_t^{(k)}\}_{k=1}^{K_t}
=
\mathrm{main\_evolution}(\hat{H}_t,h_t;M).
\end{equation}

The editable surface includes all task-facing components of the harness, such as prompts, tools, memory, state management, hooks, and execution policies. Candidate modifications are evaluated through the task-harness scope, which returns reward feedback that guides subsequent edits.

The evolution procedure itself is not prescribed by a fixed optimization schedule. Instead, $M$ determines when to inspect code, propose changes, evaluate candidates, and terminate an iteration. The only invariant preserved across all rewrites is the task-injection interface: while internal harness components may change, the external interface connecting tasks to the harness remains fixed. This invariant enables direct comparison of evolved versions and makes the harness hot-swappable across iterations.

\subsubsection{Commit Selection: Multi-Commit Pool}
\label{sec:select-commit}

After main evolution, commit selection chooses a set of candidate versions for future exploration:

\begin{equation}
C_t=
\mathrm{commit\_selection}
(\{V_t^{(k)}\}_{k=1}^{K_t},\mathcal{G}_t;M).
\end{equation}

Rather than selecting only the highest-reward candidate, HSI maintains a diverse commit pool containing multiple evolutionary directions. Each selected version is added to the evolution graph together with the semantic rationale generated by $M$, enabling future seed selection to reason over successful, failed, and unexplored branches.

\subsubsection{Meta-Evolution: Evolving the Evolver}
\label{sec:meta-evolve}

Meta-evolution extends the evolutionary process one level upward by modifying the strategy $\Sigma$ that governs harness evolution:

\begin{equation}
\Sigma_{t+1}
=
\mathrm{meta\_evolution}
(\mathcal{G}_t\cup C_t,\Sigma_t;M).
\end{equation}

The editable surface of $\Sigma$ includes the procedures responsible for seed selection, main evolution, commit selection, and final version selection. By modifying $\Sigma$, the meta-evolver changes not only candidate harnesses but also the search strategy used to discover future harnesses.

The meta-evolver is restricted to the evolution strategy space. Its own execution logic remains immutable and is loaded from the outer initialization template. This frozen anchor prevents recursive self-modification from becoming unbounded while preserving the ability to adapt the evolution process itself.

\subsubsection{Best-Version Selection: Generalization-First Export}
\label{sec:select-best}

After $T$ iterations, the final stage selects the deployed harness:

\begin{equation}
H^*
=
\mathrm{best\_version\_selection}
(\mathcal{G}_T,M,\Sigma).
\end{equation}

Unlike intermediate commit selection, which maintains diversity for future exploration, this stage prioritizes generalization. Candidate versions are evaluated on validation performance, and the selected harness is exported for held-out evaluation. The selection procedure itself remains part of $\Sigma$, allowing meta-evolution to adapt how final deployment decisions are made.

\section{Experiments}

\subsection{Experimental Setup}
\label{sec:config}

\paragraph{Benchmark and metric.}
We evaluate HSI on BALROG~\citep{paglieri2025balrog}, a benchmark of long-horizon text-based interactive environments designed to evaluate planning, memory, exploration, and tool-use capabilities of LLM agents. BALROG contains six environments with different capability demands: BabyAI, BabaIsAI, Crafter, MiniHack, TextWorld, and NLE.

BabyAI and BabaIsAI evaluate instruction following and navigation in structured environments. Crafter requires long-horizon planning, resource management, and sequential decision making. TextWorld evaluates multi-step reasoning and object manipulation. MiniHack and NLE provide increasingly challenging roguelike environments with complex state spaces and sparse feedback. Together, these environments provide a natural difficulty spectrum, ranging from tasks where the frozen backbone achieves non-trivial performance to tasks where performance remains limited.

BALROG reports performance using episode-level \% Progress~\citep{paglieri2025balrog}, which measures task completion on a $0$--$100$ scale. We rescale this metric to $[0,1]$ during evolution. Candidate harnesses are ranked using a stochastic lower-confidence-bound reward:
\begin{equation}
r=\mu-z\frac{\sigma}{\sqrt{n}},
\end{equation}
where $\mu$ and $\sigma$ denote the mean and standard deviation over evaluation trials, and $z=0.5$ in our experiments. This reward reduces the impact of stochastic high-reward trajectories during evolution. All reported results use raw \% Progress means rather than LCB rewards.

\paragraph{Benchmark rationale.}
We choose BALROG because it provides a controlled environment for studying harness evolution in long-horizon interactive tasks. Unlike static evaluation settings, BALROG requires agents to maintain state, reason over multiple steps, interact with environments through tools, and adapt behavior based on execution feedback. These characteristics expose the components that can be influenced by harness design, including memory management, state tracking, exploration strategies, and action coordination.
Furthermore, BALROG contains environments spanning different difficulty regimes within a unified evaluation framework. This allows us to study both where harness evolution provides measurable gains and where improvements remain limited under a fixed backbone, providing an empirical view of the practical boundaries of self-improving agents.

\paragraph{Backbone and evolution configuration.}
All experiments use DeepSeek-V4-Flash (accessed through the \texttt{deepseek-v4-flash-preview} API service) as the frozen backbone $M$ across the task-harness, evolver, and meta-evolver scopes. Each evolution run consists of $T=5$ outer iterations with a maximum of 80 \texttt{react()} steps per iteration.

To isolate the contribution of harness evolution from inference-time reasoning, extended reasoning is disabled in the task-harness scope and enabled for the evolver and meta-evolver scopes. This configuration remains fixed across development evaluation, validation evaluation, best-version selection, and final testing. Therefore, improvements observed during task execution cannot be attributed to additional reasoning computation at inference time.

We further constrain the evolution space such that all task interactions must be mediated through the frozen backbone $M$. Evolution may modify the harness components, including prompts, tools, memory, state management, and control logic, but cannot replace the model with external search procedures or non-LLM policies.

\paragraph{Evaluation protocols.}
BALROG environments are procedurally generated, and each \texttt{evaluate()} call samples a new initial seed. We use two evaluation protocols to measure different generalization settings.

\textit{In-distribution evolution.}
The same task set is used during evolution and final evaluation, while each evaluation episode is generated from a newly sampled environment seed. This protocol measures whether harness evolution improves performance under stochastic variations of previously encountered tasks. We apply this protocol to TextWorld, BabyAI, Crafter, MiniHack, and NLE.

During evolution, each candidate harness is evaluated with one episode per call for efficiency. For final evaluation, we use the full episode budget to obtain a more stable estimate of performance.

\textit{Held-out task generalization.}
For BabaIsAI, we construct task-family splits based on sub-suite categories: BreakStop, GoTo, Make, and Advanced. Each sub-suite is divided into development, validation, and test portions. The test split remains inaccessible throughout the evolution process.

This protocol evaluates whether an evolved harness can generalize to unseen tasks within the same task family. We report held-out results on BreakStop, GoTo, and Make. The Advanced sub-suite contains only three tasks and therefore does not provide sufficient samples for a meaningful held-out evaluation.

\paragraph{Baselines.}
We use three types of reference comparisons.
First, \textbf{Init Harness} is the original handcrafted harness evaluated without evolution under the same backbone and evaluation protocol. This serves as the primary controlled baseline for measuring the effect of harness evolution.
Second, we compare with publicly reported BALROG leaderboard results where available. These comparisons provide context against frontier models under their native configurations, including different backbones and reasoning settings.
Third, we distinguish HSI from external-proposer harness optimization approaches. Methods that rely on stronger external models to design or optimize the harness are not included in the controlled comparison because they operate under a different assumption from HSI, which studies endogenous evolution under a fixed backbone.

\subsection{HSI Performance}
\label{sec:main-results}

We evaluate HSI under the two protocols introduced in \S\ref{sec:config}. This section first studies Setup~A, where evolution and evaluation use the same task distribution with stochastic resampling, and then studies Setup~B for held-out generalization.

\subsubsection{Setup~A: In-distribution harness evolution.}

Table~\ref{tab:leaderboard} compares HSI with the BALROG leaderboard as well as the controlled init-harness baseline using the same frozen backbone. The key comparison is between the bottom three rows, where all configurations use DeepSeek-V4-Flash and differ only in whether and how the harness is evolved.

Starting from the same backbone and task-time inference configuration, HSI substantially improves over the init harness across all non-trivial suites. The meta-evolution-on configuration improves \% Progress by $+39.3$ on BabyAI, $+33.0$ on Crafter, $+25.0$ on TextWorld, and $+15.0$ on MiniHack, while keeping the backbone and task-time reasoning budget unchanged. These results indicate that a frozen LLM can obtain substantial performance gains through endogenous modification of its surrounding harness.

HSI also reaches competitive performance with frontier systems under their native configurations. On TextWorld, HSI achieves $65.0$ \% Progress, exceeding Grok-4 ($62.9$), Claude-Opus-4.5-Thinking ($59.0$), and Gemini-3-Flash ($50.2$). On Crafter, HSI reaches $44.6$, outperforming DeepSeek-R1 ($36.4$), GPT-5-minimal-think ($39.1$), and GPT-4o ($33.1$). These comparisons are provided as contextual references, while the controlled init-harness comparison isolates the effect of harness evolution.

\begin{table}[t]
\caption{BALROG leaderboard comparison  under \emph{Setup~A}. The top block lists public-leaderboard numbers (retrieved 2026-08-03) from frontier models under their native configurations, reported as \% Progress~\citep{paglieri2025balrog} (mean $\pm$ standard deviation across evaluation episodes). The bottom block reports HSI on the same task suite using a single frozen DeepSeek-V4-Flash backbone, isolating the contribution of the continuously evolvable hot-swappable task harness: the init-harness baseline, the meta-evolution-off arm, and the \textbf{meta-evolution-on arm (bolded)} that exports the deployed harness. BabaIsAI is omitted because our sub-suite protocol (\S\ref{sec:cross-env}) differs from the leaderboard's mixed-task protocol. \textbf{Avg} reports the unweighted mean across the five environments, with its std the unweighted mean of the per-environment stds.}
\label{tab:leaderboard}
\centering
\footnotesize
\setlength{\tabcolsep}{2pt}
\begin{tabular}{lcccccc}
\toprule
\textbf{LLM} & \textbf{BabyAI} & \textbf{Crafter} & \textbf{TextWorld} & \textbf{MiniHack} & \textbf{NLE} & \textbf{Avg} \\
\midrule
Gemini-3-Pro              & $96.0 \pm 2.8$ & $57.3 \pm 4.4$ & $60.2 \pm 7.5$ & $40.0 \pm 7.7$ & $6.8 \pm 3.2$ & $52.1 \pm 5.1$ \\
Gemini-3.1-Pro-Thinking   & $98.0 \pm 2.0$ & $55.0 \pm 6.4$ & $75.7 \pm 6.4$ & $27.5 \pm 7.1$ & $2.6 \pm 0.3$ & $51.8 \pm 4.4$ \\
Gemini-3.1-Pro            & $100.0 \pm 0.0$ & $46.8 \pm 4.2$ & $66.5 \pm 7.5$ & $35.0 \pm 7.5$ & $3.0 \pm 0.5$ & $50.3 \pm 3.9$ \\
Gemini-3-Flash            & $86.0 \pm 4.9$ & $45.0 \pm 6.3$ & $50.2 \pm 8.1$ & $30.0 \pm 7.2$ & $4.0 \pm 0.8$ & $43.0 \pm 5.5$ \\
Grok-4                    & $76.0 \pm 6.0$ & $57.3 \pm 3.9$ & $62.9 \pm 7.9$ & $17.5 \pm 6.0$ & $1.8 \pm 0.8$ & $43.1 \pm 4.9$ \\
Claude-Opus-4.5           & $80.0 \pm 5.7$ & $49.5 \pm 3.1$ & $51.4 \pm 8.4$ & $27.5 \pm 7.1$ & $2.0 \pm 0.5$ & $42.1 \pm 5.0$ \\
Claude-Opus-4.5-Thinking  & $72.0 \pm 6.3$ & $48.6 \pm 3.2$ & $59.0 \pm 8.0$ & $30.0 \pm 7.2$ & $2.4 \pm 0.3$ & $42.4 \pm 5.0$ \\
Gemini-2.5-Pro-Exp-03-25  & $80.0 \pm 5.7$ & $55.0 \pm 6.0$ & $49.2 \pm 8.2$ & $17.5 \pm 6.0$ & $1.7 \pm 0.2$ & $40.7 \pm 5.2$ \\
DeepSeek-R1               & $74.0 \pm 6.2$ & $36.4 \pm 3.8$ & $21.8 \pm 6.1$ & $25.0 \pm 6.8$ & $1.4 \pm 0.5$ & $31.7 \pm 4.7$ \\
GPT-5-minimal-think       & $80.0 \pm 5.7$ & $39.1 \pm 4.1$ & $30.6 \pm 7.0$ & $20.0 \pm 7.3$ & $1.3 \pm 0.5$ & $34.2 \pm 4.9$ \\
Claude-3.5-Sonnet         & $68.0 \pm 6.6$ & $32.7 \pm 3.2$ & $42.1 \pm 5.4$ & $15.0 \pm 5.6$ & $0.6 \pm 0.5$ & $31.7 \pm 4.3$ \\
GPT-4o                    & $77.6 \pm 3.7$ & $33.1 \pm 2.3$ & $39.3 \pm 5.2$ & $10.0 \pm 4.7$ & $0.4 \pm 0.4$ & $32.1 \pm 3.3$ \\
\midrule
DS-V4-Flash (Init harness) & $42.0 \pm 3.5$ & $11.6 \pm 5.0$ & $40.0 \pm 6.2$ & $0.8 \pm 1.9$ & $0.0$ & $18.9 \pm 3.3$ \\
\midrule
DS-V4-Flash w. HSI (meta-off)        & $77.3 \pm 1.2$ & $36.4 \pm 1.6$ & $46.0 \pm 2.4$ & $5.8 \pm 3.8$ & $0.0$ & $33.1 \pm 1.8$ \\
DS-V4-Flash w. HSI (meta-on)         & {\boldmath $81.3 \pm 4.2$} & {\boldmath $44.6 \pm 3.2$} & {\boldmath $65.0 \pm 3.0$} & {\boldmath $15.8 \pm 2.9$} & {\boldmath $0.2 \pm 0.3$} & {\boldmath $41.4 \pm 2.7$} \\
\bottomrule
\end{tabular}
\end{table}

\paragraph{Effect of meta-evolution.}

The meta-off ablation in Table~\ref{tab:leaderboard} isolates the contribution of evolving the evolution strategy itself. Removing the meta-evolver reduces performance on every evaluated suite: BabyAI decreases from $81.3$ to $77.3$, Crafter from $44.6$ to $36.4$, TextWorld from $65.0$ to $46.0$, and MiniHack from $15.8$ to $5.8$. The largest improvements occur on TextWorld ($+19.0$) and MiniHack ($+10.0$), suggesting that adapting the evolution procedure becomes increasingly beneficial as the harness search space becomes more complex.

NLE does not provide a meaningful meta-off comparison because both configurations achieve near-zero reward. The meta-on result of $0.2$ indicates that evolution receives insufficient task feedback to discover useful harness modifications.

\paragraph{Evolution dynamics and capability boundary.}

Across evolution runs, the largest performance improvement typically occurs during the first iteration, followed by smaller incremental gains in later iterations. This behavior suggests that early harness redesign captures the dominant improvements, while later iterations refine existing solutions through exploration and selection.

The performance pattern across BALROG also reveals a practical limitation of harness evolution. While HSI improves tasks where the frozen backbone produces informative feedback, it does not substantially improve NLE, where the initial capability and reward signal are both extremely limited. This observation is consistent with the feedback and capability constraints discussed in \S\ref{sec:design}: harness evolution can reorganize the behavior surrounding a model, but cannot overcome tasks where the model cannot generate useful interaction signals.

\subsubsection{Setup~B: held-out generalization on BabaIsAI sub-suites.}
\label{sec:cross-env}

To evaluate generalization beyond the evolution tasks, we perform held-out evaluation on three BabaIsAI sub-suites: \textbf{BreakStop}, \textbf{GoTo}, and \textbf{Make}. Each sub-suite evolves an independent harness while keeping the backbone, evolution budget, initialization template, and evaluation protocol fixed. The only varying factor is the task family. All experiments use the same frozen DeepSeek-V4-Flash backbone with the task-time no-thinking configuration described in \S\ref{sec:config}, and compare meta-evolution-on and meta-evolution-off variants.

Each sub-suite follows Setup~B with a $20\%$ held-out test split. The Advanced sub-suite is excluded because it contains only three tasks, which is insufficient for a meaningful split evaluation. Table~\ref{tab:babaisai} summarizes the results.

\begin{table}[t]
\caption{BabaIsAI sub-suite results under \emph{Setup~B} (split evolution with $20\%$ held-out test). ``Best Dev'' denotes the highest development reward selected during evolution. Test results are reported as mean $\pm$ across-task standard deviation of task progress. Init Harness is averaged over three baseline runs.}
\label{tab:babaisai}
\centering
\small
\setlength{\tabcolsep}{4pt}
\begin{tabular}{lcccc}
\toprule
\textbf{Sub-suite} & \textbf{Init Harness} & \textbf{Best Dev} & \textbf{Best Test (meta-on)} & \textbf{Best Test (meta-off)} \\
\midrule
BreakStop & $0.0333 \pm 0.0334$ & 1.0000 & $0.9800 \pm 0.0632$ & $1.0000 \pm 0.0000$ \\
GoTo      & $0.1818 \pm 0.0802$ & 1.0000 & $1.0000 \pm 0.0000$ & $0.9636 \pm 0.0809$ \\
Make      & $0.0000$             & 0.5556 & $0.3625 \pm 0.3284$ & $0.3375 \pm 0.2029$ \\
\bottomrule
\end{tabular}
\end{table}

The results show two distinct regimes. For navigation-oriented tasks (BreakStop and GoTo), HSI achieves near-perfect held-out performance: the meta-on variant reaches $0.98$ and $1.00$ test progress respectively, while the meta-off variant achieves comparable results. These results indicate that the evolved harness discovers reusable interaction patterns that transfer beyond the observed development tasks.

In contrast, Make remains substantially more challenging. Although harness evolution improves over the zero-shot init harness, held-out performance remains limited ($0.36$ for meta-on and $0.34$ for meta-off). The smaller improvement and larger variance suggest that multi-step crafting requires capabilities beyond the reusable harness transformations discovered during evolution.

Across BALROG environments, the same qualitative pattern emerges: harness evolution provides the largest improvements on tasks where the frozen backbone already exhibits meaningful competence, smaller and noisier gains on tasks near the capability boundary, and limited improvement on tasks where useful feedback is difficult to obtain. This observation is consistent with the two practical limitations discussed in \S\ref{sec:design}: harness evolution can reorganize and amplify existing model capabilities, but cannot fully overcome insufficient feedback or fundamental backbone limitations.

\subsubsection{Evolution trajectory analysis.}
\begin{figure}[t]
\centering
\includegraphics[width=\linewidth]{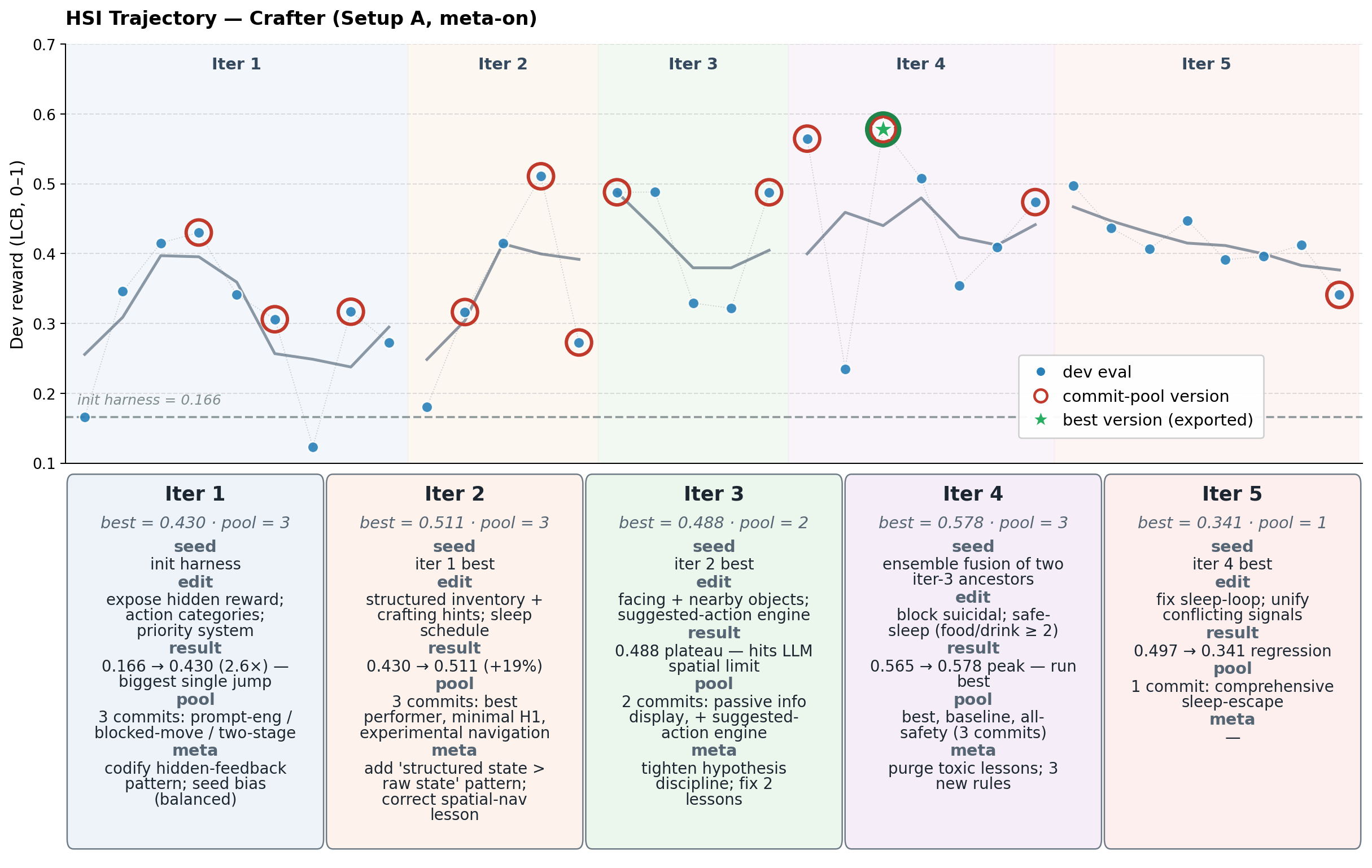}
\caption{HSI evolution trajectory on Crafter (Setup~A, meta-on). The dev reward climbs from the init-harness baseline $0.166$ to an iteration-best $0.578$ by iteration~4 (green-starred as the exported best version), with a regression in iteration~5. Each iteration's annotation card reports four fields on the task-harness and evolver scopes (seed origin, main-evolution edit, result, commit pool) plus a green-tinted \emph{meta} field summarizing what the meta-evolver rewrote in the evolver strategy $\Sigma$ that iteration. The dominant lever uncovered by the evolver is \emph{making hidden game feedback explicit}: reward signal, inventory state, and crafting feasibility, successively exposed in the harness context; the meta-evolver's contribution is to codify these patterns into $\Sigma$ so later iterations inherit them.}
\label{fig:trajectory-crafter}
\end{figure}

\begin{figure}[t]
\centering
\includegraphics[width=\linewidth]{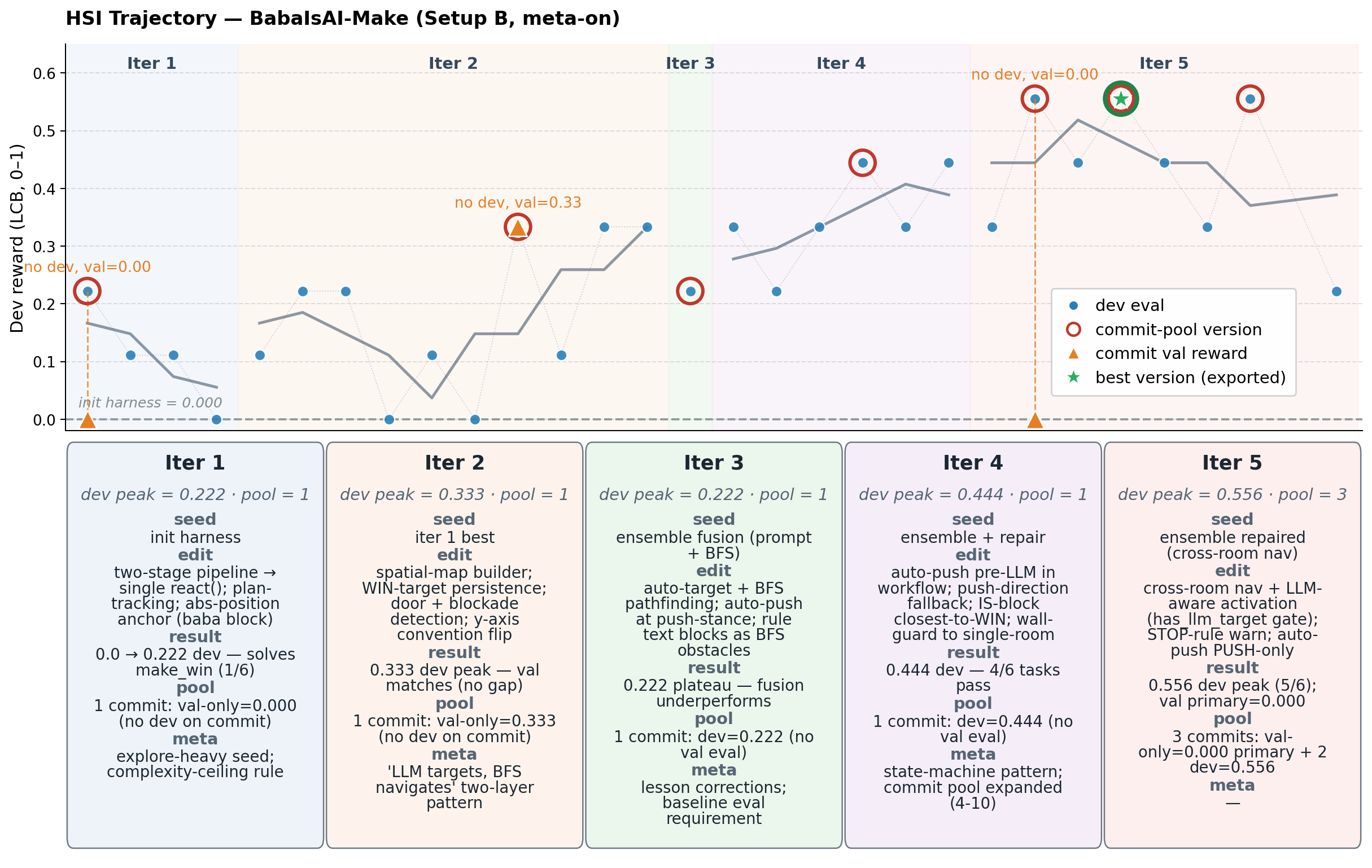}
\caption{HSI evolution trajectory on BabaIsAI-Make (Setup~B, meta-on). The main blue curve is the dev reward across all \texttt{reward\_history} evaluations; red rings mark commit-pool versions. Orange triangles ($\triangle$) flag commits finalized in val mode (no dev evaluation recorded on the commit itself): each $\triangle$ sits at the commit's anchor $x$ with the val reward on the y-axis, and a dashed vertical line drops from the dev anchor to the val reward so the dev$\to$val evaluation structure reads directly off the trajectory. The dev peak climbs $0.222 \to 0.333 \to 0.444 \to 0.556$ across five iterations as the agent introduces a single-react rewrite with plan tracking (iter~1), a spatial-map builder with WIN-target persistence (iter~2), auto-target computation with BFS pathfinding (iter~3), an auto-push mechanism with directional fallback (iter~4), and LLM-aware cross-room navigation (iter~5); the meta-evolver codifies the ``LLM targets, BFS navigates'' two-layer pattern and progressively expands the commit pool from one to three versions. The $\triangle$ markers expose which iterations' primary commits were validated on held-out val tasks, an evaluation structure unique to Setup~B.}
\label{fig:trajectory-make}
\end{figure}

 To understand how self-improvement emerges, we visualize representative evolution trajectories from Crafter (Setup~A, Figure~\ref{fig:trajectory-crafter}) and BabaIsAI-Make (Setup~B, Figure~\ref{fig:trajectory-make}). Each trajectory records the reward evolution across five iterations together with the selected seeds, harness modifications, commit pools, and meta-evolution updates.

\paragraph{Crafter trajectory.} The Crafter trajectory illustrates the typical in-distribution evolution pattern. The first iterations primarily introduce missing task representations: exposing latent reward signals, structuring inventory information, and improving action-state alignment. Later iterations explore more specialized mechanisms, including rule-based suggestions and safety constraints. The best version appears at iteration~4, after which additional modifications produce regression, demonstrating that evolution does not monotonically improve every branch but instead searches a non-convex harness design space.

The meta-evolver contributes by transforming successful local discoveries into reusable evolution heuristics. Across iterations, it updates $\Sigma$ with higher-level principles such as prioritizing structured state representations over raw observations and avoiding overly aggressive exploration near a performance plateau. These changes affect future search behavior rather than directly modifying task performance.

\paragraph{BabaIsAI-Make trajectory.}  The BabaIsAI-Make trajectory demonstrates held-out evolution. Unlike Crafter, where evaluation is performed under stochastic resampling of the same task set, Setup~B explicitly exposes validation behavior. The dev reward improves through successive harness redesigns, including spatial abstraction, BFS-based planning, and LLM-guided navigation. The held-out validation markers show that several discovered mechanisms transfer beyond the development tasks, although the remaining gap indicates that multi-step crafting still approaches the capability boundary of the frozen backbone.

Across both trajectories, HSI exhibits a common evolution pattern: initial iterations discover missing abstractions, middle iterations introduce structured algorithmic components, and later iterations refine or prune competing designs. This provides qualitative evidence that self-improvement occurs through progressive harness restructuring rather than simple prompt optimization.

\section{Discussion and Conclusion}

\subsection{Lessons from Hierarchical Self-Improvement}

Running HSI across BALROG environments reveals several observations about when endogenous harness evolution is effective.

\begin{itemize}

\item \textbf{Execution feedback is the foundation of reliable evolution.}

Harness modifications are difficult to evaluate from static inspection alone. The useful signal is whether the modified harness produces improved behavior when executed in the environment. This observation motivates HSI's design choices: maintaining a fixed task-injection interface, evaluating candidate versions through interaction rather than code-level heuristics, and using reward feedback to guide successive modifications. Without execution-grounded feedback, self-modification can easily produce plausible-looking changes that do not translate into improved task performance.

\item \textbf{Single-seed evolution improves attribution clarity.}

Harness evolution is inherently stochastic: different search trajectories can discover different improvements. HSI intentionally avoids population-based parallel scaling and instead follows a single evolving lineage, making it possible to attribute performance changes to harness redesign rather than to increased candidate throughput. This design trades search efficiency for clearer measurement of endogenous improvement. Population-based exploration~\citep{zhang2026darwingodelmachineopenended,weng2026groupevolvingagentsopenendedselfimprovement} remains complementary and could be incorporated as an additional scaling dimension.

\item \textbf{Task-specific evolution appears to be a practical scaling direction.}

The results suggest that evolved harnesses capture task-family-specific structure rather than universally transferable solutions. In BabaIsAI, strong held-out performance on BreakStop and GoTo demonstrates that evolution can discover reusable strategies within a task family, while the same mechanisms do not automatically transfer across substantially different environments. This indicates that future scaling of self-improving agents may benefit from maintaining specialized evolvable harnesses for different task distributions rather than relying solely on a single universal harness.

\item \textbf{Backbone capability determines the reachable improvement frontier.}

Harness evolution improves performance when the environment provides informative feedback and the frozen model can effectively exploit that information. Across our experiments, large gains occur on tasks within the model's reachable competence range (TextWorld, BabyAI, Crafter, and BabaIsAI-GoTo/BreakStop), while harder environments exhibit smaller or negligible improvements (MiniHack, BabaIsAI-Make, and NLE). These results suggest an empirical capability boundary: harness evolution expands the effectiveness of a fixed model, but cannot completely overcome limitations in the model's underlying reasoning ability or insufficient environmental feedback.

\end{itemize}

\subsection{Conclusion}

HSI presents a hierarchical self-improvement framework in which a single frozen LLM $M$ operates across three scopes: the task harness that interacts with the environment, the evolver that rewrites the harness, and the meta-evolver that rewrites the evolution strategy itself. By separating these scopes through editable boundaries and preserving a frozen outer anchor, HSI enables recursive improvement while avoiding unrestricted self-reference.

Experiments on BALROG demonstrate that a fixed DeepSeek-V4-Flash backbone can substantially improve through endogenous harness evolution alone. Under in-distribution evaluation, HSI consistently improves over the initial harness on multiple environments while keeping the backbone and task-time inference configuration unchanged. Under held-out evaluation on BabaIsAI, evolved harnesses generalize to unseen tasks within the same task family, demonstrating that the improvements are not limited to individual trajectories.

At the same time, the results reveal clear limits. Evolution requires informative feedback, and environments with extremely sparse rewards provide insufficient signals for improvement. Moreover, the final performance remains constrained by the capability of the underlying frozen model. These observations suggest that self-improving agents should be viewed not as replacing stronger models, but as a mechanism for systematically extracting additional capability from existing models through environment-grounded harness adaptation.

HSI provides one instantiation of this paradigm: a per-task, continuously evolvable harness framework under a frozen backbone. Future directions include combining hierarchical evolution with stronger foundation models, richer feedback environments, and scalable population-based search.

\section*{Acknowledgments}

This work was developed as an open-source research project. The authors acknowledge the support of the open-source communities and the developers of the models, environments, and tools that made this project possible.

The current release represents an initial exploration of hierarchical self-improvement under a frozen language model. Due to practical computational constraints, the evaluation focuses on a selected set of benchmarks, backbones, and comparisons rather than a full-scale empirical study. We view this release as a foundation for continued community exploration, including evaluations on additional models, environments, and self-improvement settings.

The complete source code is available at:
\url{https://github.com/TailinZhou/hsi}.
We welcome researchers and practitioners to build upon, reproduce, and extend the framework.
%%%%%%%%%%%%%%%%%%%%%%%%%%%%%%%%%%%%%%%%%%%%%%%%%%%%%%%%%%%%

%%%%%%%%%%%%%%%%%%%%%%%%%%%%%%%%%%%%%%%%%%%%%%%%%%%%%%%%%%%%

\bibliographystyle{plainnat}
\bibliography{references}

@misc{xin2026eurekagentagentenvironmentengineering,
      title={EurekAgent: Agent Environment Engineering is All You Need For Autonomous Scientific Discovery}, 
      author={Amy Xin and Jiening Siow and Junjie Wang and Zijun Yao and Fanjin Zhang and Jian Song and Lei Hou and Juanzi Li},
      year={2026},
      eprint={2606.13662},
      archivePrefix={arXiv},
      primaryClass={cs.AI},
      url={https://arxiv.org/abs/2606.13662}, 
}

@misc{chen2026harnessxcomposableadaptiveevolvable,
      title={HarnessX: A Composable, Adaptive, and Evolvable Agent Harness Foundry}, 
      author={Tingyang Chen and Shuo Lu and Kang Zhao and Weicheng Meng and Hanlin Teng and Tianhao Li and Chao Li and Xule Liu and Jian Liang and Zhizhong Zhang and Yuan Xie and Heng Qu and Kun Shao and Jian Luan},
      year={2026},
      eprint={2606.14249},
      archivePrefix={arXiv},
      primaryClass={cs.AI},
      url={https://arxiv.org/abs/2606.14249}, 
}

@misc{lee2026recursiveharnessselfimprovement,
      title={Recursive Harness Self-Improvement}, 
      author={Hyunin Lee and Jinglue Xu and Jeffrey Seely and Donghyun Lee and Matei Zaharia and Yujin Tang},
      year={2026},
      eprint={2607.15524},
      archivePrefix={arXiv},
      primaryClass={cs.LG},
      url={https://arxiv.org/abs/2607.15524}, 
}

@misc{luo2026selfevolvingagentharnessesgated,
      title={Self-Evolving Agent Harnesses via Gated Semantic Quality-Diversity}, 
      author={Xiaotian Luo and Fengxingyu Wang and Chuanrui Hu and Dizhan Xue and Yafeng Deng},
      year={2026},
      eprint={2607.13683},
      archivePrefix={arXiv},
      primaryClass={cs.CL},
      url={https://arxiv.org/abs/2607.13683}, 
}

@misc{wang2026rethinkingevaluationharnessevolution,
      title={Rethinking the Evaluation of Harness Evolution for Agents}, 
      author={Yike Wang and Huaisheng Zhu and Zhengyu Hu and Yige Yuan and Zhengyu Chen and Shakti Senthil and Hannaneh Hajishirzi and Yulia Tsvetkov and Pradeep Dasigi and Teng Xiao},
      year={2026},
      eprint={2607.12227},
      archivePrefix={arXiv},
      primaryClass={cs.AI},
      url={https://arxiv.org/abs/2607.12227}, 
}

@misc{yang2026recursivemultiagentsystems,
      title={Recursive Multi-Agent Systems}, 
      author={Xiyuan Yang and Jiaru Zou and Rui Pan and Ruizhong Qiu and Pan Lu and Shizhe Diao and Jindong Jiang and Hanghang Tong and Tong Zhang and Markus J. Buehler and Jingrui He and James Zou},
      year={2026},
      eprint={2604.25917},
      archivePrefix={arXiv},
      primaryClass={cs.AI},
      url={https://arxiv.org/abs/2604.25917}, 
}

@misc{nie2026tthetesttimeharnessevolution,
      title={TTHE: Test-Time Harness Evolution}, 
      author={Jun Nie and Yonggang Zhang and Jun Song and Qianshu Cai and Dahai Yu and Yike Guo and Xinmei Tian and Bo Han},
      year={2026},
      eprint={2607.08124},
      archivePrefix={arXiv},
      primaryClass={cs.SE},
      url={https://arxiv.org/abs/2607.08124}, 
}

@misc{luo2026harnessawareselfevolvingcoevolvingmodel,
      title={Harness-Aware Self-Evolving: Co-Evolving Model Weights, Harness, and Task Solutions}, 
      author={Haochen Luo and Yi Huang and Sichun Luo and Fengyuan Liu and Lei Li and Zefa Hu and Junlan Feng and Qi Liu},
      year={2026},
      eprint={2607.03935},
      archivePrefix={arXiv},
      primaryClass={cs.AI},
      url={https://arxiv.org/abs/2607.03935}, 
}

@misc{shen2026skilloptlitebetterfasteragent,
      title={SkillOpt-Lite: Better and Faster Agent Self-evolution via One Line of Vibe}, 
      author={Yifei Shen and Bo Li and Xinjie Zhang},
      year={2026},
      eprint={2607.03451},
      archivePrefix={arXiv},
      primaryClass={cs.SE},
      url={https://arxiv.org/abs/2607.03451}, 
}

@misc{zheng2026seagymevaluationenvironmentselfevolving,
      title={SEAGym: An Evaluation Environment for Self-Evolving LLM Agents}, 
      author={Congjie Zheng and Chuanyi Xue and Bin Liang and Jun Yang and Changshui Zhang},
      year={2026},
      eprint={2606.17546},
      archivePrefix={arXiv},
      primaryClass={cs.AI},
      url={https://arxiv.org/abs/2606.17546}, 
}

@misc{lumer2026recursiveagentharnesses,
      title={Recursive Agent Harnesses}, 
      author={Elias Lumer and Sahil Sen and Kevin Paul and Vamse Kumar Subbiah},
      year={2026},
      eprint={2606.13643},
      archivePrefix={arXiv},
      primaryClass={cs.CL},
      url={https://arxiv.org/abs/2606.13643}, 
}

@misc{zhang2026selfharnessharnessesimprove,
      title={Self-Harness: Harnesses That Improve Themselves}, 
      author={Hangfan Zhang and Shao Zhang and Kangcong Li and Chen Zhang and Yang Chen and Yiqun Zhang and Lei Bai and Shuyue Hu},
      year={2026},
      eprint={2606.09498},
      archivePrefix={arXiv},
      primaryClass={cs.CL},
      url={https://arxiv.org/abs/2606.09498}, 
}

@misc{liu2026adaptiveautoharnesssustainedselfimprovement,
      title={Adaptive Auto-Harness: Sustained Self-Improvement for Agentic System Deployment on Open-Ended Task Streams}, 
      author={Zewen Liu and Zhan Shi and Yisi Sang and Bing He and Minhua Lin and Tianxin Wei and Dakuo Wang and Benoit Dumoulin and Wei Jin and Hanqing Lu},
      year={2026},
      eprint={2606.01770},
      archivePrefix={arXiv},
      primaryClass={cs.LG},
      url={https://arxiv.org/abs/2606.01770}, 
}

@misc{chen2026harnessforgejointharnesspolicy,
      title={HarnessForge: Joint Harness and Policy Evolution for Adaptive Agent Systems}, 
      author={Mingju Chen and Can Lv and Guibin Zhang and Heng Chang and Shiji Zhou},
      year={2026},
      eprint={2606.01779},
      archivePrefix={arXiv},
      primaryClass={cs.CL},
      url={https://arxiv.org/abs/2606.01779}, 
}

@misc{lin2026harnessupdatingharnessbenefit,
      title={Harness Updating Is Not Harness Benefit: Disentangling Evolution Capabilities in Self-Evolving LLM Agents}, 
      author={Minhua Lin and Juncheng Wu and Zijun Wang and Zhan Shi and Yisi Sang and Bing He and Zewen Liu and Tianxin Wei and Zongyu Wu and Zhiwei Zhang and Dakuo Wang and Xiang Zhang and Benoit Dumoulin and Cihang Xie and Yuyin Zhou and Suhang Wang and Hanqing Lu},
      year={2026},
      eprint={2605.30621},
      archivePrefix={arXiv},
      primaryClass={cs.AI},
      url={https://arxiv.org/abs/2605.30621}, 
}

@misc{hebbar2026siaselfimprovingai,
      title={SIA: Self Improving AI with Harness \& Weight Updates},
      author={Prannay Hebbar and Yogendra Manawat and Samuel Verboomen and Alesia Ivanova and Selvam Palanimalai and Kunal Bhatia and Vignesh Baskaran},
      year={2026},
      eprint={2605.27276},
      archivePrefix={arXiv},
      primaryClass={cs.AI},
      url={https://arxiv.org/abs/2605.27276}, 
}

@misc{xu2026adaptinginterfacemodelruntime,
      title={Adapting the Interface, Not the Model: Runtime Harness Adaptation for Deterministic LLM Agents}, 
      author={Tianshi Xu and Huifeng Wen and Meng Li},
      year={2026},
      eprint={2605.22166},
      archivePrefix={arXiv},
      primaryClass={cs.AI},
      url={https://arxiv.org/abs/2605.22166}, 
}

@misc{yao2026harnessbenchmeasuringharnesseffects,
      title={Harness-Bench: Measuring Harness Effects across Models in Realistic Agent Workflows}, 
      author={Yilun Yao and Xinyu Tan and Chao-Hsuan Liu and Yaoming Li and Zhengyang Wang and Wenhan Yu and Zhewen Tan and Yuxuan Tian and Guangxiang Zhao and Lin Sun and Xiangzheng Zhang and Tong Yang},
      year={2026},
      eprint={2605.27922},
      archivePrefix={arXiv},
      primaryClass={cs.AI},
      url={https://arxiv.org/abs/2605.27922}, 
}

@misc{yang2026skilloptexecutivestrategyselfevolving,
      title={SkillOpt: Executive Strategy for Self-Evolving Agent Skills}, 
      author={Yifan Yang and Ziyang Gong and Weiquan Huang and Qihao Yang and Ziwei Zhou and Zisu Huang and Yan Li and Xuemei Gao and Qi Dai and Bei Liu and Kai Qiu and Yuqing Yang and Dongdong Chen and Xue Yang and Chong Luo},
      year={2026},
      eprint={2605.23904},
      archivePrefix={arXiv},
      primaryClass={cs.AI},
      url={https://arxiv.org/abs/2605.23904}, 
}

@misc{lin2026agenticharnessengineeringobservabilitydriven,
      title={Agentic Harness Engineering: Observability-Driven Automatic Evolution of Coding-Agent Harnesses}, 
      author={Jiahang Lin and Shichun Liu and Chengjun Pan and Lizhi Lin and Shihan Dou and Zhiheng Xi and Xuanjing Huang and Hang Yan and Zhenhua Han and Tao Gui and Yu-Gang Jiang},
      year={2026},
      eprint={2604.25850},
      archivePrefix={arXiv},
      primaryClass={cs.CL},
      url={https://arxiv.org/abs/2604.25850}, 
}

@misc{karten2026continualharnessonlineadaptation,
      title={Continual Harness: Online Adaptation for Self-Improving Foundation Agents}, 
      author={Seth Karten and Joel Zhang and Tersoo Upaa Jr and Ruirong Feng and Wenzhe Li and Chengshuai Shi and Chi Jin and Kiran Vodrahalli},
      year={2026},
      eprint={2605.09998},
      archivePrefix={arXiv},
      primaryClass={cs.LG},
      url={https://arxiv.org/abs/2605.09998}, 
}

@misc{wei2026architecturaldesigndecisionsai,
      title={Architectural Design Decisions in AI Agent Harnesses}, 
      author={Hu Wei},
      year={2026},
      eprint={2604.18071},
      archivePrefix={arXiv},
      primaryClass={cs.AI},
      url={https://arxiv.org/abs/2604.18071}, 
}

@misc{lee2026metaharnessendtoendoptimizationmodel,
      title={Meta-Harness: End-to-End Optimization of Model Harnesses}, 
      author={Yoonho Lee and Roshen Nair and Qizheng Zhang and Kangwook Lee and Omar Khattab and Chelsea Finn},
      year={2026},
      eprint={2603.28052},
      archivePrefix={arXiv},
      primaryClass={cs.AI},
      url={https://arxiv.org/abs/2603.28052}, 
}

@misc{zhang2026darwingodelmachineopenended,
      title={Darwin Godel Machine: Open-Ended Evolution of Self-Improving Agents}, 
      author={Jenny Zhang and Shengran Hu and Cong Lu and Robert Lange and Jeff Clune},
      year={2026},
      eprint={2505.22954},
      archivePrefix={arXiv},
      primaryClass={cs.AI},
      url={https://arxiv.org/abs/2505.22954}, 
}

@misc{lou2026autoharnessimprovingllmagents,
      title={AutoHarness: improving LLM agents by automatically synthesizing a code harness}, 
      author={Xinghua Lou and Miguel Lázaro-Gredilla and Antoine Dedieu and Carter Wendelken and Wolfgang Lehrach and Kevin P. Murphy},
      year={2026},
      eprint={2603.03329},
      archivePrefix={arXiv},
      primaryClass={cs.CL},
      url={https://arxiv.org/abs/2603.03329}, 
}

@misc{weng2026groupevolvingagentsopenendedselfimprovement,
      title={Group-Evolving Agents: Open-Ended Self-Improvement via Experience Sharing}, 
      author={Zhaotian Weng and Antonis Antoniades and Deepak Nathani and Zhen Zhang and Xiao Pu and Xin Eric Wang},
      year={2026},
      eprint={2602.04837},
      archivePrefix={arXiv},
      primaryClass={cs.AI},
      url={https://arxiv.org/abs/2602.04837}, 
}

@misc{wang2026statisticallimitsselfimprovingagents,
      title={On The Statistical Limits of Self-Improving Agents}, 
      author={Charles L. Wang and Keir Dorchen and Peter Jin},
      year={2026},
      eprint={2510.04399},
      archivePrefix={arXiv},
      primaryClass={cs.AI},
      url={https://arxiv.org/abs/2510.04399}, 
}

@misc{xia2025livesweagentsoftwareengineeringagents,
      title={Live-SWE-agent: Can Software Engineering Agents Self-Evolve on the Fly?}, 
      author={Chunqiu Steven Xia and Zhe Wang and Yan Yang and Yuxiang Wei and Lingming Zhang},
      year={2025},
      eprint={2511.13646},
      archivePrefix={arXiv},
      primaryClass={cs.SE},
      url={https://arxiv.org/abs/2511.13646}, 
}

@misc{wang2025huxleygodelmachinehumanlevelcoding,
      title={Huxley-G\"odel Machine: Human-Level Coding Agent Development by an Approximation of the Optimal Self-Improving Machine}, 
      author={Wenyi Wang and Piotr Piekos and Li Nanbo and Firas Laakom and Yimeng Chen and Mateusz Ostaszewski and Mingchen Zhuge and J{\"u}rgen Schmidhuber},
      year={2025},
      eprint={2510.21614},
      archivePrefix={arXiv},
      primaryClass={cs.AI},
      url={https://arxiv.org/abs/2510.21614}, 
}

@misc{yin2025godelagentselfreferentialagent,
      title={G\"odel Agent: A Self-Referential Agent Framework for Recursive Self-Improvement}, 
      author={Xunjian Yin and Xinyi Wang and Liangming Pan and Li Lin and Xiaojun Wan and William Yang Wang},
      year={2025},
      eprint={2410.04444},
      archivePrefix={arXiv},
      primaryClass={cs.AI},
      url={https://arxiv.org/abs/2410.04444}, 
}

@misc{robeyns2025selfimprovingcodingagent,
      title={A Self-Improving Coding Agent},
      author={Maxime Robeyns and Martin Szummer and Laurence Aitchison},
      year={2025},
      eprint={2504.15228},
      archivePrefix={arXiv},
      primaryClass={cs.AI},
      url={https://arxiv.org/abs/2504.15228},
}

@misc{wu2026automemautomatedlearningmemory,
      title={AutoMem: Automated Learning of Memory for LLM Agents},
      author={Wu and others},
      year={2026},
      note={Cited for context on memory as a cognitive skill component; excluded from the harness evolution survey},
}

@misc{schmidhuber2003godelmachine,
      title={G\"{o}del Machines: Fully Self-Referential Universal Self-Improvers},
      author={J{\"u}rgen Schmidhuber},
      year={2003},
      eprint={cs/0309048},
      archivePrefix={arXiv},
      primaryClass={cs.LG},
      url={https://arxiv.org/abs/cs/0309048},
}

@misc{zhang2026hyperagents,
      title={HyperAgents: Metacognitive Self-Improvement via Editable Meta-Mechanisms},
      author={Jenny Zhang and Shengran Hu and Cong Lu and Robert Lange and Jeff Clune},
      year={2026},
      eprint={2603.19461},
      archivePrefix={arXiv},
      primaryClass={cs.AI},
      url={https://arxiv.org/abs/2603.19461},
}

@inproceedings{paglieri2025balrog,
    title={{BALROG}: Benchmarking Agentic {LLM} and {VLM} Reasoning On Games},
    author={Davide Paglieri and Bart{\l}omiej Cupia{\l} and Samuel Coward and Ulyana Piterbarg and Maciej Wolczyk and Akbir Khan and Eduardo Pignatelli and {\L}ukasz Kuci{\'n}ski and Lerrel Pinto and Rob Fergus and Jakob Nicolaus Foerster and Jack Parker-Holder and Tim Rockt{\"a}schel},
    booktitle={The Thirteenth International Conference on Learning Representations},
    year={2025},
    url={https://openreview.net/forum?id=fp6t3F669F}
}

@misc{weng2026harness,
      title={Harness Engineering for Self-Improvement},
      author={Lilian Weng},
      year={2026},
      howpublished={\url{https://lilianweng.github.io/posts/2026-07-04-harness/}},
}
\appendix

\section{Implementation Details of the HSI Agent System}
\label{app:implementation}

This section provides additional implementation details of HSI that are omitted from the main text. The goal is to describe the execution environment, tool interfaces, memory organization, and scope isolation mechanisms that realize the hierarchical architecture.

\subsection{Agent Execution Interface}
\label{app:agent-interface}

All HSI components are instantiated using the same frozen LLM $M$ and a shared \texttt{react()} execution primitive. Each scope provides a different tool set and system context, while the underlying model remains unchanged.

At each step, the model receives the current message history, available tools, and task-specific context, and produces an action through the following loop:

\begin{equation}
a_t = M(o_t, \mathcal{T}, \mathcal{C}_t),
\end{equation}

where $o_t$ denotes the current observation, $\mathcal{T}$ is the available tool set, and $\mathcal{C}_t$ represents the scope-specific context. The selected action may modify files, request evaluation, record information, or terminate the current stage.

The three scopes differ only through their tool availability:

\begin{itemize}
    \item \textbf{Task-harness scope:} interacts with the benchmark environment and executes the current harness $H$.
    \item \textbf{Evolver scope:} modifies files within the harness directory and invokes evaluation on development tasks.
    \item \textbf{Meta-evolver scope:} modifies files within the evolution strategy directory \texttt{evolution/}.
\end{itemize}

This separation ensures that the same model can operate at different abstraction levels without introducing additional learned parameters or external optimizer models.

\subsection{Evolution Tool Interface}
\label{app:tools}

The evolver scope is equipped with a small set of atomic tools that allow $M$ to inspect, modify, and evaluate candidate harnesses.

The file manipulation interface includes:

\begin{itemize}
    \item \texttt{read}: inspect existing source files;
    \item \texttt{write}: create new files;
    \item \texttt{edit}: modify existing implementations;
    \item \texttt{bash}: execute shell commands for debugging or verification.
\end{itemize}

In addition to file operations, HSI provides evolution-specific primitives:

\begin{itemize}
    \item \texttt{plan}: maintain an iteration-local reasoning notebook;
    \item \texttt{compact\_context}: summarize previous interactions when the context budget becomes limiting;
    \item \texttt{evaluate}: execute the current harness and return environment feedback;
    \item \texttt{lesson}: record reusable insights for future iterations;
    \item \texttt{end\_evolution}: terminate the current evolution process.
\end{itemize}

The framework does not specify an ordering among these operations. The model decides when and how to use each tool according to the observed feedback and current evolution objective.

\subsection{Memory Organization}
\label{app:memory}

HSI maintains multiple memory channels with different persistence properties.

\paragraph{Iteration-local memory.}

During a single evolution iteration, the evolver maintains a temporary notebook stored in \texttt{plan.md}. This memory stores intermediate hypotheses, debugging notes, and short-term decisions. If the candidate version is discarded, this memory is rolled back together with the corresponding code state.

\paragraph{Persistent evolutionary memory.}

Across iterations, HSI maintains a persistent lesson archive stored in \texttt{BOOTSTRAP.md}. Entries in this archive summarize previously discovered patterns, failed directions, and reusable evolution guidance. Seed selection can access this information when generating hypotheses for future iterations.

\paragraph{Evolution graph memory.}

The cumulative evolution graph $\mathcal{G}_t$ stores committed harness versions and their relationships. Each node contains:

\begin{itemize}
    \item the corresponding harness snapshot;
    \item the achieved reward;
    \item metadata describing the evolution step.
\end{itemize}

Edges encode semantic relationships between versions, such as extending an existing approach, repairing a failure mode, or exploring a different direction. This graph provides long-term structural memory for future seed selection.

\subsection{Trajectory Compression and Probe Mechanism}
\label{app:probe}

Directly exposing all historical trajectories to the LLM would exceed the available context budget and introduce unnecessary noise. HSI therefore uses a probe mechanism that retrieves compressed summaries from previous execution histories.

The probe operates as an auxiliary querying process:

\begin{equation}
z = \mathrm{probe}(\mathcal{T}_{history}, q),
\end{equation}

where $\mathcal{T}_{history}$ represents stored trajectories and $q$ is the query specified by the current agent. The probe returns a compact summary rather than raw interaction traces.

For example, the meta-evolver may query:

\begin{itemize}
    \item which seed-selection behaviors correlate with successful iterations;
    \item which evolution patterns frequently lead to regression;
    \item which hypothesis structures precede large improvements.
\end{itemize}

This mechanism allows the meta-evolver to exploit long-horizon evolutionary history while keeping its reasoning context bounded.

\subsection{Scope Isolation and Editable Boundaries}
\label{app:scope}

HSI enforces explicit boundaries between the three hierarchical scopes.

The task harness $H$ is editable by the evolver but cannot modify the evolution strategy $\Sigma$. Conversely, the meta-evolver can modify $\Sigma$ but cannot directly alter the task harness during meta-evolution. Any modification outside the authorized directory is rejected.

The execution logic of the meta-evolver itself is loaded from an immutable initialization template. Therefore, the recursive modification process terminates at a fixed outer boundary:

\begin{equation}
M
\rightarrow
H
\rightarrow
\Sigma
\rightarrow
\text{frozen anchor}.
\end{equation}

This hierarchy provides the structural constraint required for controlled self-modification: lower layers remain adaptable, while the outermost execution boundary remains fixed.

\subsection{Evaluation Interface}
\label{app:evaluation-interface}

Evaluation is performed outside the editable surfaces. The harness receives tasks through a fixed interface:

\begin{equation}
\texttt{using\_harness(agent, task)}.
\end{equation}

Although the internal implementation of $H$ may change across iterations, this interface remains invariant. Therefore, all evolved versions operate under the same task injection mechanism and can be compared using identical development, validation, and test protocols.

The evaluator returns both a scalar reward and optional textual feedback. The scalar reward is used for candidate comparison, while textual feedback provides qualitative information that can guide subsequent evolution.

\section{Technical appendices and supplementary material}

\subsection{Per-Suite Experimental Configuration}
\label{sec:per-suite-config}

Table~\ref{tab:per-suite-config} records the experimental configuration for each BALROG suite used in this paper. The shared configuration across all suites is as follows. The backbone is DeepSeek-V4-Flash accessed via the \texttt{deepseek-v4-flash-preview} API service. The evolver scope runs with $T=5$ outer iterations, at most $80$ \texttt{react()} steps per iteration, LCB reward at $z=0.5$, thinking enabled, and reasoning effort at maximum. The task-harness scope runs with thinking disabled and temperature $0$. The meta-evolver scope runs with at most $50$ \texttt{react()} steps per iteration, a greedy archive, seed selection and commit pooling both evolvable, the seed hypothesis injected into each iteration's first system prompt, and a short seed-validation probe (up to three \texttt{evaluate()} calls) enabled during seed selection. The terminal best-version selection stage is a fixed, non-evolvable agentic stage that runs once at the end of every evolution. The init harness is not pre-evaluated; iteration~1 starts cold.

Only the entries in Table~\ref{tab:per-suite-config} vary across suites. ``Setup'' refers to the two protocols defined in \S\ref{sec:config} (A: full-set in-distribution, B: sub-suite split with held-out test). ``Dev ratio'' and ``Val ratio'' are the fractions of the suite assigned to dev (evolution reward signal) and val (best-version selection). ``Test episodes'' is the per-task episode count at the final test evaluation. ``Dev episodes'' is the per-task episode count during evolution (cheaper dev-time estimation, with noise absorbed by the LCB reward and the final test measurement). ``Test repeats'' is the number of times the full test set is re-evaluated after evolution. ``Meta'' indicates whether the meta-evolver scope is enabled. ``Submit-best steps'' is the step budget of the terminal best-version selection stage.

\begin{table}[h]
\caption{Per-suite experimental configuration. Shared settings are listed in the prose above; only the entries below vary across suites.}
\label{tab:per-suite-config}
\centering
\small
\begin{tabular}{lcccccccc}
\toprule
\textbf{Suite} & \textbf{Setup} & \textbf{Dev} & \textbf{Val} & \textbf{Test ep.} & \textbf{Dev ep.} & \textbf{Test rep.} & \textbf{Meta} & \textbf{Submit-best} \\
\midrule
TextWorld           & A & 1.0 & 0.00 & 10 & 3 & 3 & off & 50 \\
BabyAI              & A & 1.0 & 0.00 & 10 & 3 & 3 & on  & 80 \\
Crafter             & A & 1.0 & 0.00 & 5  & 3 & 3 & on  & 50 \\
MiniHack            & A & 1.0 & 0.00 & 5  & 1 & 3 & on  & 80 \\
NLE                 & A & 1.0 & 0.00 & 5  & 1 & 1 & on  & 50 \\
BabaIsAI-BreakStop  & B & 0.8 & 0.20 & 5  & 1 & 1 & off & 80 \\
BabaIsAI-GoTo       & B & 0.8 & 0.25 & 5  & 1 & 1 & on  & 80 \\
BabaIsAI-Make       & B & 0.8 & 0.25 & 5  & 1 & 1 & on  & 80 \\
\bottomrule
\end{tabular}
\end{table}

\subsection{Detailed Survey of Harness Engineering and Agent Self-Evolution Methods}

We provide an extended survey of 31 representative works in harness engineering and agent self-evolution (2025--2026), organized by research direction. AutoMem~\citep{wu2026automemautomatedlearningmemory} is excluded from this survey as it focuses on memory as a cognitive skill component rather than harness evolution per se.

\subsubsection{G\"{o}del-Style Recursive Self-Improvement}

G\"{o}del Agent~\citep{yin2025godelagentselfreferentialagent} (Yin et al., Peking Univ./UCSB, 2025). The first LLM-based self-referential agent framework inspired by the G\"{o}del Machine. Implements self-referential improvement through recursive main function: self-inspect (read current algorithm), interact (evaluate performance), self-update (alter code via LLM-generated monkey patches), continue-improve (recursively invoke decision algorithm). Key results: constrained version (G\"{o}del-base, GPT-3.5) outperforms all baselines: DROP 80.9\%, MGSM 64.2\%, MMLU 70.9\%, GPQA 34.9\%; full evolution $\approx\$$15 (vs.\ Meta Agent Search's \$$300$); on Game of 24, agent spontaneously switches from LLM methods to search algorithms achieving 100\% accuracy.

Darwin G\"{o}del Machine (DGM)~\citep{zhang2026darwingodelmachineopenended} (Zhang et al., UBC/Vector Institute/Sakana AI, ICLR 2026). The first practical FM-based G\"{o}del Machine combining self-referential code modification with population-based open-ended exploration. Initialized with a single lightweight coding agent (Claude 3.5 Sonnet, only bash + edit tools). Over 80 iterations: parent selected from archive proportional to performance and inversely proportional to number of children; parent analyzes benchmark logs, proposes new features, modifies its own codebase (Turing-complete Python); staged evaluation (10 $\to$ 50 $\to$ 200 tasks). Key results: SWE-bench 20.0\% $\to$ 50.0\%; Polyglot 14.2\% $\to$ 30.7\% (surpassing handcrafted Aider); greedy ablation only 39.7\% vs.\ full DGM 50.0\%, confirming archive-based exploration is essential; automatically discovered granular file editing, multi-patch ranking, auto-summarization, and history-aware patch generation; archive tree reveals key innovations triggered via lower-performing intermediate nodes.

Huxley-G\"{o}del Machine (HGM)~\citep{wang2025huxleygodelmachinehumanlevelcoding} (Wang et al., KAUST, 2025). Identifies and resolves the mismatch between benchmark performance and self-improvement potential. Introduces Clade-level Meta-Productivity (CMP): aggregating descendant performance across an agent's entire clade to measure self-improvement potential. Uses Thompson sampling over Beta distributions of clade success rates for node expansion. Theorem 1 proves that under reasonable assumptions, access to true CMP suffices to simulate the G\"{o}del Machine. Key results: SWE-Verified-60 56.7\% (vs.\ DGM 53.3\%, SICA 50.0\%); Polyglot 30.5\% (vs.\ DGM 27.1\%); 2.38$\times$--6.86$\times$ faster than DGM; CMP estimates correlate with empirical CMP at 0.778 vs.\ DGM's 0.285; agent discovered iterative refinement and nested diff-patch structures; transfers to GPT-5 achieves 57.0\% on SWE-Lite, matching best handcrafted agents.

Group-Evolving Agents (GEA)~\citep{weng2026groupevolvingagentsopenendedselfimprovement} (Weng et al., UC Santa Barbara, 2026). Shifts the evolutionary unit from individuals to groups with explicit intra-group experience sharing. At each iteration, $K{=}2$ parents are selected via Performance-Novelty scoring; their evolutionary traces (code patches, predicted task patches, execution logs, evaluation outcomes) are aggregated into a shared pool; each agent's reflection module (GPT-o1) analyzes shared experience to generate evolution directives. Key results: SWE-bench Verified 71.0\% (vs.\ DGM 56.7\%); Polyglot 88.3\% (vs.\ DGM 68.3\%); best agent integrates 8/9 key innovations from 17 unique ancestors (28.3\% of population) vs.\ DGM's 5/9 from 9 ancestors (15.0\%); worst top-5 GEA agent (58.3\%) strictly beats DGM's best (56.7\%); framework-level bugs repaired in 1.4 iterations (vs.\ 5.0 for DGM).

SICA (Self-Improving Coding Agent)~\citep{robeyns2025selfimprovingcodingagent} (Robeyns et al., Univ.\ of Bristol/iGent AI, 2025). The first fully self-referential coding agent that collapses the distinction between meta-agent and target-agent: the agent edits its own full Python codebase. Self-improvement loop: evaluate on benchmarks $\to$ best agent promoted to meta-agent $\to$ archive analysis sub-agent proposes improvement $\to$ software-developer sub-agent implements $\to$ independent verification. Key results: SWE-bench (50-problem subset) 17\% $\to$ 53\% ($+36$pp, 15 iterations); agent autonomously invented Smart Editor, AST Symbol Locator, Hybrid Symbol Locator, context-sensitive diff minimization. Critical negative result: Agent Framework Saturation. On AIME 2024/GPQA Diamond reasoning tasks, scaffolding offered \emph{no gain} over the raw model (plateau at $\sim$76\% vs.\ bare o3-mini-high at 87\%), suggesting that for strong reasoning models, scaffolding may \emph{interrupt} rather than augment internal reasoning chains.

HyperAgents~\citep{zhang2026hyperagents} (Zhang et al., UBC/Vector Institute/Edinburgh/NYU/FAIR at Meta/Meta Superintelligence Labs, 2026). Extends DGM by making the meta-mechanism itself editable. Instead of DGM's hand-designed, fixed instruction-generation function, the task agent and meta agent are fused into a single editable Python program (the \emph{hyperagent}), so the procedure that generates new agents is itself a target of self-modification. Key results: Polyglot (training subset) 0.140 $\to$ 0.340 (CI 0.300--0.380), matching DGM's coding performance without coding-specific hand-design; paper review 0.0 $\to$ 0.710, exceeding the open-source AI-Scientist-v2 static baseline (0.630); robot reward design 0.060 $\to$ 0.372, exceeding the direct-reward default (0.348); cross-domain transfer from paper review $+$ robot reward to IMO grading yields imp@50 $=$ 0.630 vs.\ DGM-custom's $\approx$ 0, with the combined pipeline (transfer $+$ 200 iterations) reaching 0.640. Two ablations (without self-improvement; without open-ended exploration) yield near-zero improvement on paper review and robot reward design ($p < 0.05$), confirming both components are necessary. Qualitative finding: hyperagents spontaneously evolve persistent memory and performance-tracking infrastructure, domain-agnostic meta-level capabilities that emerge rather than being specified.

Recursive Agent Harnesses (RAH)~\citep{lumer2026recursiveagentharnesses} (Lumer et al., PwC, 2026). Uses the full agent harness (filesystem tools, code execution, planning) as the recursive unit rather than bare model calls. Parent agent spawns subagents via code-execution spawning (Python scripts with \texttt{asyncio.gather}, bypassing per-turn API caps) or JSON tool-call spawning. Key results: GPT-5 backbone fixed, RAH improves Codex baseline from 71.75\% to 81.36\% ($+9.61$pp); sonnet-4.5 + RAH reaches 89.77\%; robust across 1K--4M token context lengths.

EurekAgent~\citep{xin2026eurekagentagentenvironmentengineering} (Xin et al., 2026). Frames ``environment engineering'' as a core research direction, designing agent environments along four dimensions: permissions engineering (Docker isolation, hidden evaluators), artifact engineering (filesystem + Git shared memory), budget engineering, and human-in-the-loop engineering. Key results: new SOTA on three math tasks and TriMul CUDA kernel engineering (4.3\% improvement over best human); MLE-Bench Lite 85.71\% any-medal rate; 26-circle packing SOTA discovered with \$\,$<$\,11 API cost.

\subsubsection{Harness Engineering and Online Adaptation}

Meta-Harness~\citep{lee2026metaharnessendtoendoptimizationmodel} (Lee et al., Stanford/KRAFTON/MIT, 2026). The seminal work that formalizes harness optimization as outer-loop code-space search. A stronger coding agent (Claude Code + Opus-4.6) serves as the proposer with full filesystem access to all prior harness candidates' source code, scores, and execution traces, up to 10M tokens of diagnostic information per evaluation, three orders of magnitude larger than prior text optimizers (OPRO, TextGrad, OpenEvolve). The proposer selectively inspects prior artifacts via \texttt{grep}, \texttt{cat}, etc. (median 82 files per iteration, referencing $>20$ prior candidates). Key results: $+7.7$pp over ACE on online text classification with $4\times$ fewer context tokens; $+4.7$pp average improvement across 5 held-out models on IMO-level math reasoning; \#1 Haiku-4.5 agent on TerminalBench-2 (37.6\%). The proposer discovered environment bootstrapping strategies that human experts had not explicitly designed.

AutoHarness~\citep{lou2026autoharnessimprovingllmagents} (Lou et al., Google DeepMind, 2026). Demonstrates that Gemini-2.5-Flash can synthesize code harnesses via Thompson-sampling-guided tree search. Two harness templates: harness-as-action-verifier (rejection sampling with learned \texttt{is\_legal\_action()}) and harness-as-policy (code directly outputs actions without LLM calls). Key results: 100\% legal action rate on 145 TextArena games (vs.\ 78\% of Gemini losses from illegal moves in Kaggle chess); harness-as-policy achieves 0.870 average reward, surpassing larger models.

Architectural Design Decisions in AI Agent Harnesses~\citep{wei2026architecturaldesigndecisionsai} (Wei, 2026). A protocol-guided empirical study of 70 publicly available agent projects identifying five recurring design dimensions (subagent architecture, context management, tool systems, safety mechanisms, orchestration) with their option spectra. Reports co-occurrences (e.g., execution isolation with structured safety control at lift 3.4) and five archetypal architectural patterns.

Self-Harness~\citep{zhang2026selfharnessharnessesimprove} (Zhang et al., Shanghai AI Lab, 2026). Proposes that a fixed model can improve its own operating harness without external help, operationalized as three stages: Weakness Mining (clusters execution traces by verifier-grounded failure signatures), Harness Proposal (generates diverse minimal candidate modifications), and Proposal Validation (non-regression acceptance rule on held-in and held-out splits). Key results: MiniMax M2.5 40.5\% $\to$ 61.9\% ($+53\%$ relative); Qwen3.5-35B-A3B 23.8\% $\to$ 38.1\% ($+60\%$); GLM-5 42.9\% $\to$ 57.1\% ($+33\%$). Qualitative analysis revealed model-specific adaptations: different models produced completely different harness modifications from the same initial harness and algorithm.

Agentic Harness Engineering (AHE)~\citep{lin2026agenticharnessengineeringobservabilitydriven} (Lin et al., Fudan/Peking Univ., 2026). Argues that the bottleneck in harness evolution is observability, not agent capability. Three pillars: Component Observability (every editable component exposed as a file), Experience Observability (millions of raw trajectory tokens distilled into layered, drill-down evidence corpus), and Decision Observability (every edit paired with a self-declared prediction later verified). Key results: TerminalBench-2 pass@1 from 69.7\% to 77.0\% in 10 iterations; frozen harness transfers to SWE-bench-verified with highest aggregate success while using 12\% fewer tokens. Self-attribution: fix-precision 33.7\%, fix-recall 51.4\% ($5\times$ random), but regression-precision only 11.8\% ($2\times$ random).

HarnessX~\citep{chen2026harnessxcomposableadaptiveevolvable} (Darwin Agent Team, 2026). Three innovations: (1) compositional harness formalized as typed processors attached to lifecycle hooks with a nine-dimensional taxonomy; (2) AEGIS evolution engine built on an ``operational mirror'' between symbolic adaptation and RL theory with defenses against reward hacking, catastrophic forgetting, and under-exploration; (3) harness-model co-evolution via cross-harness GRPO. Key results: average $+14.5\%$ gain across 15 model-benchmark configurations ($+44.0\%$ on ALFWorld with Qwen3.5-9B); variant isolation resolves stagnation on heterogeneous GAIA ($\Delta=0.0 \to +13.6\%$); co-evolution adds $+4.7\%$ over harness-only. All three predicted RL pathologies confirmed empirically.

HarnessForge~\citep{chen2026harnessforgejointharnesspolicy} (Chen et al., Beihang/Tsinghua, 2026). Formalizes LLM agent systems as harness-policy pairs $G=(H,R)$ where $H=(P,A,M)$ specifies Planning, Action, and Memory components. Iteratively alternates fault-guided harness tailoring with harness-conditioned policy alignment (supervised trace alignment using successful trajectories). Key results: $+3.56\%$ average over strongest per-metric baselines on ToolHop/SearchQA/TMDB/API-Bank; compatibility matrix shows best harness with mismatched policy underperforms, proving harness engineering alone insufficient.

TTHE~\citep{nie2026tthetesttimeharnessevolution} (Nie et al., HKBU/USTC, 2026). Treats the executable harness as the state of test-time adaptation, maintaining a population of candidate harnesses and refining them using only unlabeled execution traces. Key results: BIRD 12.0\% $\to$ 50.0\%; LiveCodeBench 30.0\% $\to$ 38.3\%; SWE-bench 20.0\% $\to$ 35.0\%; claw-eval 48.9\% $\to$ 69.8\% ($+20.9$pp). Oracle analysis reveals $\sim$14pp selection regret and $\sim$30\% coverage gap, identifying proxy signal reliability as the central challenge.

Live-SWE-Agent~\citep{xia2025livesweagentsoftwareengineeringagents} (Xia et al., UIUC, 2025). The first ``live'' software agent that self-evolves at runtime by creating, modifying, and using custom tools during problem-solving. Starts from a minimal scaffold (mini-SWE-agent, $\sim$100 LOC, bash only) and adds a reflection step after each action. Key results: SWE-bench Verified 77.4\% with Gemini 3 Pro, outperforming all known agents; zero offline cost, $\$$0.02--0.12 overhead per task; critical finding: GPT-5-Nano \emph{degrades} ($-68.2\%$ relative) when attempting to self-evolve, establishing a model capability floor for self-evolution.

Continual Harness~\citep{karten2026continualharnessonlineadaptation} (Karten et al., Princeton/Google DeepMind, 2026). Extends harness evolution to embodied agents in reset-free online settings. Refiner edits full harness state (system prompt, sub-agents, skills, memory) every $F$ steps within a single continuous episode. Key results: Gemini Plays Pokemon, first AI to complete multiple Pokemon RPGs; from-scratch Continual Harness recovers majority of gap to hand-engineered expert harness; Flash-Lite stalls below 20\%, indicating a capability floor.

Adaptive Auto-Harness~\citep{liu2026adaptiveautoharnesssustainedselfimprovement} (Liu et al., Emory/Amazon, 2026). Formalizes harness construction as regret minimization with decomposition into evolution loss $L_{\text{evo}}$ and adaptation loss $L_{\text{adapt}}$. Key results: PolyBench 80.9\% accuracy (vs.\ Meta-Harness 50.8\%); CTF-Dojo 50.2\% Pass@1; adaptation loss remains positive across all cycles, confirming a single committed harness is never sufficient for streaming heterogeneous workloads.

LIFE-HARNESS~\citep{xu2026adaptinginterfacemodelruntime} (Xu et al., Peking Univ., 2026). Organizes harness adaptation into four lifecycle layers: Environment Contract, Procedural Skill, Action Realization, and Trajectory Regulation. Key results: improves 116/126 (92\%) model-environment settings with 88.5\% average relative improvement; harness evolved from Qwen3-4B transfers to 17 other models.

\subsubsection{Harness-Model Co-Evolution}

HASE~\citep{luo2026harnessawareselfevolvingcoevolvingmodel} (Luo et al., HKU/China Mobile, 2026). Places solution generation and harness editing into a unified GRPO action space, enabling a single model to co-evolve weights, harness, and task solutions. Distinguishes guidance components (freely editable) from evaluation components (only fixable upon local vs.\ real-world evaluator divergence). Key results: Qwen3-8B matches GPT-OSS-120B + Claude Code on text classification (86.98\% vs.\ 86.8\%); Alpha factor mining IR 1.70 (vs.\ GPT-OSS-120B's 0.80); progressive evaluator repair in circle packing.

SIA~\citep{hebbar2026siaselfimprovingai} (Hebbar et al., Hexo Labs/Oxford, 2026). First system updating both harness and model weights in a single closed-loop. Feedback-Agent dynamically selects between harness updates and weight updates (PPO/entropic advantage weighting/GRPO). Key results: LawBench 70.1\% (vs.\ harness-only 50.0\%); TriMul CUDA kernel 1,017$\mu$s, 14.02$\times$ speedup over harness-only; harness and weight updates operate in complementary change spaces.

SkillOpt \& SkillOpt-Lite~\citep{yang2026skilloptexecutivestrategyselfevolving, shen2026skilloptlitebetterfasteragent} (Yang et al., Microsoft; Shen et al., 2026). SkillOpt introduces deep-learning-style controls to text-space skill optimization (rollout batches, textual learning rate, validation gate, rejected-edit buffer, slow/meta update). SkillOpt-Lite formalizes this as Zeroth-Order optimization with three principles (consensus mining, validation gating, filesystem-based exploration). HarnessOpt enables a smaller model with optimized harness to surpass larger models.

\subsubsection{Recursive and Multi-Agent Harness Systems}

Recursive Harness Self-Improvement (RHI)~\citep{lee2026recursiveharnessselfimprovement} (Lee et al., Sakana AI/UC Berkeley, 2026). Defines harness as a prompt-level specification (roles, instructions, contracts, hops) and iteratively refines it using trajectory-local self-comparison (pairwise preference feedback against the immediately previous harness only, not a population). Key results: sonnet-4.6-high + 2 RHI iterations outperforms sonnet-4.6-max; opus-4.8-high + RHI outperforms opus-4.8-ultracode; inference cost reduced up to 60\%; gains arise from improved communication contracts reducing cross-component redundancy.

Recursive Multi-Agent Systems (RecursiveMAS)~\citep{yang2026recursivemultiagentsystems} (Yang et al., Stanford/UIUC/NVIDIA/MIT, 2026). Extends recursive scaling from single models to multi-agent systems via latent-space recursive computation with lightweight RecursiveLink modules. Key results: 8.3\% average accuracy improvement across 9 benchmarks; 1.2$\times$--2.4$\times$ inference speedup; 34.6\%--75.6\% token reduction; only 13.12M trainable parameters (0.31\% of full model).

\subsubsection{Evaluation, Attribution, and Theoretical Foundations}

Rethinking the Evaluation of Harness Evolution~\citep{wang2026rethinkingevaluationharnessevolution} (Wang et al., AI2/UW, 2026). The pivotal critique of harness evolution research. Controlled experiments matching feedback and inference budgets show: without unit tests, harness evolution 67.4\% vs.\ parallel sampling 72.3\%; with unit tests, 75.8\% vs.\ 86.0\%; on disjoint search/evaluation tasks, only $+0.6$pp generalization gain. Concludes that most harness edits ``memorize fixes rather than distilling strategies'' and remaining failures stem from model limitations, not harness deficiencies. Calls for benchmarks satisfying two conditions: (1) tasks are sufficiently hard with room for improvement; (2) performance highly depends on specialized tools/skills/workflows. This directly motivates our BALROG-based experimental design.

Harness Updating Is Not Harness Benefit~\citep{lin2026harnessupdatingharnessbenefit} (Lin et al., 2026). First systematic decoupling of harness-updating and harness-benefit across 7 LLMs $\times$ 3 benchmarks. Key findings: harness-updating is flat in base capability (Qwen3.5-9B matches Claude Opus 4.6); harness-benefit is non-monotonic, with mid-tier models benefiting most ($+19.3$pp), strong-tier hitting ceiling ($+2.6$pp), and weak-tier benefiting least due to low skill-load rates (25.1\% vs.\ 95.7\%) and low adherence (35.0\% vs.\ 75.7\%). This provides direct empirical precedent for our NLE negative result.

GSME (Gated Semantic Quality-Diversity)~\citep{luo2026selfevolvingagentharnessesgated} (Luo et al., EverMind AI, 2026). Establishes the gold standard for trustworthy harness evaluation. Three gates: validity gate (infrastructure failures re-run), activation gate (patch credited only if it actually fired), significance gate (paired $2\sigma$ test, $z \geq 1.96$). GSME archive keyed on (defect location $\times$ defect cause) pathology rather than tasks fixed, an anti-overfitting inductive bias. Key results: sealed-test gains $+9$ to $+15.5$pp across 7 domains, retaining 86--147\% of training gain. A non-statistical ``mean improves'' rule credits $\sim$60\% of truly neutral mechanisms as wins.

Harness-Bench~\citep{yao2026harnessbenchmeasuringharnesseffects} (Yao et al., Peking Univ., 2026). First diagnostic benchmark with harness as primary evaluation axis: 106 sandboxed offline tasks, 8 workflow categories, 6 configurable harnesses $\times$ 8 model backends. Key findings: 23.8-point gap between best and worst harness; stronger models show lower cross-harness variance; five recurring failure modes mapped to harness intervention points.

SEAGym~\citep{zheng2026seagymevaluationenvironmentselfevolving} (Zheng et al., Tsinghua, 2026). RL-style evaluation environment converting Harbor-compatible benchmarks into dynamic self-evolution task sources with multi-view evaluation (update-validation, ID/OOD transfer, replay diagnostics, cost records). Key findings: batch size is non-monotonic (batch 20 optimal); source diversity critical for recovery from intermediate collapses; useful intermediate states can collapse and recover later, invisible from final scores alone.

On The Statistical Limits of Self-Improving Agents~\citep{wang2026statisticallimitsselfimprovingagents} (Wang et al., Columbia Univ., 2025). Establishes precise ``if and only if'' statistical learning theory boundaries for self-modifying agents. Core result: distribution-free PAC guarantees are preserved under self-modification \emph{iff} the policy-reachable hypothesis family has uniformly bounded VC dimension (Theorem 1). Two-Gate guardrail (validation margin + capacity cap) yields finite-sample safety with standard VC-rate oracle inequalities (Theorem 2). Identifies the Utility-Learning Tension: utility-driven changes that improve immediate performance can erode statistical preconditions for reliable generalization. Under Unbounded Representational Power, there exists a reasonable utility and PAC-learnable problem such that proof-triggered edits render the problem distribution-free unlearnable. A five-axis decomposition (Algorithmic, Representational, Architectural, Substrate, Metacognitive) unifies all forms of self-modification under a single capacity criterion. This provides the \emph{theoretical foundation} for our empirical finding that harness evolution cannot transcend model capability boundaries: when the task-required function complexity exceeds the model's reachable VC dimension, no amount of harness engineering can close the gap.

\subsection{Comparative Summary}
\begin{table}[h]
\caption{Representative harness evolution and self-improvement methods compared with HSI. \emph{Proposer}: who proposes harness edits; \emph{Surface}: the editable code surface; \emph{Domain}: primary evaluation domain; \emph{Feature}: the method's distinguishing contribution in short form.}
\label{tab:comparison}
\centering
\footnotesize
\setlength{\tabcolsep}{4pt}
\renewcommand{\arraystretch}{1.15}
\begin{tabular}{@{}p{2.5cm} p{1.9cm} p{2.0cm} p{1.8cm} p{3.0cm}@{}}
\toprule
\textbf{Method} & \textbf{Proposer} & \textbf{Surface} & \textbf{Domain} & \textbf{Feature} \\
\midrule
Meta-Harness      & External stronger  & Full harness code  & Coding, math      & Full-trajectory feedback \\
Self-Harness      & Self (target)      & Config interface   & Coding            & Model-specific edits \\
AHE               & External stronger  & Decoupled comps    & Coding            & Observability bottleneck \\
HarnessX          & Multi-agent        & Typed processors   & 5 benchmarks      & Operational mirror to RL \\
DGM               & Self               & Full codebase      & Coding            & Archive-based search \\
HGM               & Self               & Full codebase      & Coding            & Clade meta-productivity \\
GEA               & Self (group)       & Codebase + exp     & Coding            & Shared experience pool \\
SICA              & Self               & Full codebase      & Coding            & Framework saturation \\
HyperAgents       & Fused self         & Codebase + meta    & Coding, robots    & Editable meta-mechanism \\
Live-SWE-Agent    & Self (runtime)     & Tools on-the-fly   & Coding            & Zero offline cost \\
TTHE              & Self (test-time)   & Harness population & Coding, SQL       & Unlabeled trace adaptation \\
Rethinking Eval.  & --- (critique)     & ---                & Coding            & Test-time-scaling confound \\
Statistical Limits & --- (theory)      & ---                & PAC learning      & VC bound $\iff$ learnability \\
\midrule
\textbf{HSI (Ours)} & \textbf{Same frozen $M$} & \textbf{3-layer hierarchy} & \textbf{BALROG} & \textbf{Endogenous hierarchy with frozen outer anchor} \\
\bottomrule
\end{tabular}
\end{table}

\end{document}